\pdfoutput=1
\documentclass[10pt]{article}
\pdfoutput=1

\usepackage[utf8]{inputenc}
\usepackage[T1]{fontenc}
\usepackage[usenames,dvipsnames]{color}
\usepackage[protrusion=true,expansion=true,final,babel]{microtype}

\usepackage{fullpage}

\usepackage[colorlinks,linkcolor=red,anchorcolor=blue,citecolor=blue,urlcolor=blue]{hyperref}
\usepackage[numbers]{natbib}

\usepackage[subcaption]{styles/smile}
\usepackage{styles/cme-math}

\usepackage{booktabs}
\usepackage{tabularx}
\usepackage{longtable}
\usepackage{array}
\usepackage{enumitem}
\usepackage{listings}
\usepackage{graphicx}
\usepackage{tikz}
\usetikzlibrary{arrows.meta,positioning,calc,shapes.geometric,decorations.pathreplacing,fit}
\usepackage{titling}
\usepackage{cleveref}
\crefname{appendix}{Appendix}{Appendices}
\Crefname{appendix}{Appendix}{Appendices}
\crefname{subappendix}{Appendix}{Appendices}
\Crefname{subappendix}{Appendix}{Appendices}

\usepackage[left]{lineno}

\newcommand{\lean}[1]{\texttt{#1}}               

\newif\ifhasbib
\hasbibtrue

\title{L3M: Capability Without Authority\\
  \large A Verifiable Harness and Runtime for Self-Extending and Self-Improving Agents}
\author{Carson Eisenach\thanks{Correspondence to \texttt{carson@eisenach.org}.}, Robert Joseph George,
  Vincent Quenneville-B\'elair, Kate Gwimm, Dean Foster\\[3pt]
  \normalsize Amazon}
\date{September 30, 2026}

\begin{document}
\maketitle



\begin{nolinenumbers}
\noindent
\begin{minipage}{\textwidth}
  \captionsetup{type=figure,font=small,skip=4pt}
  \centering
  \includegraphics[width=\linewidth]{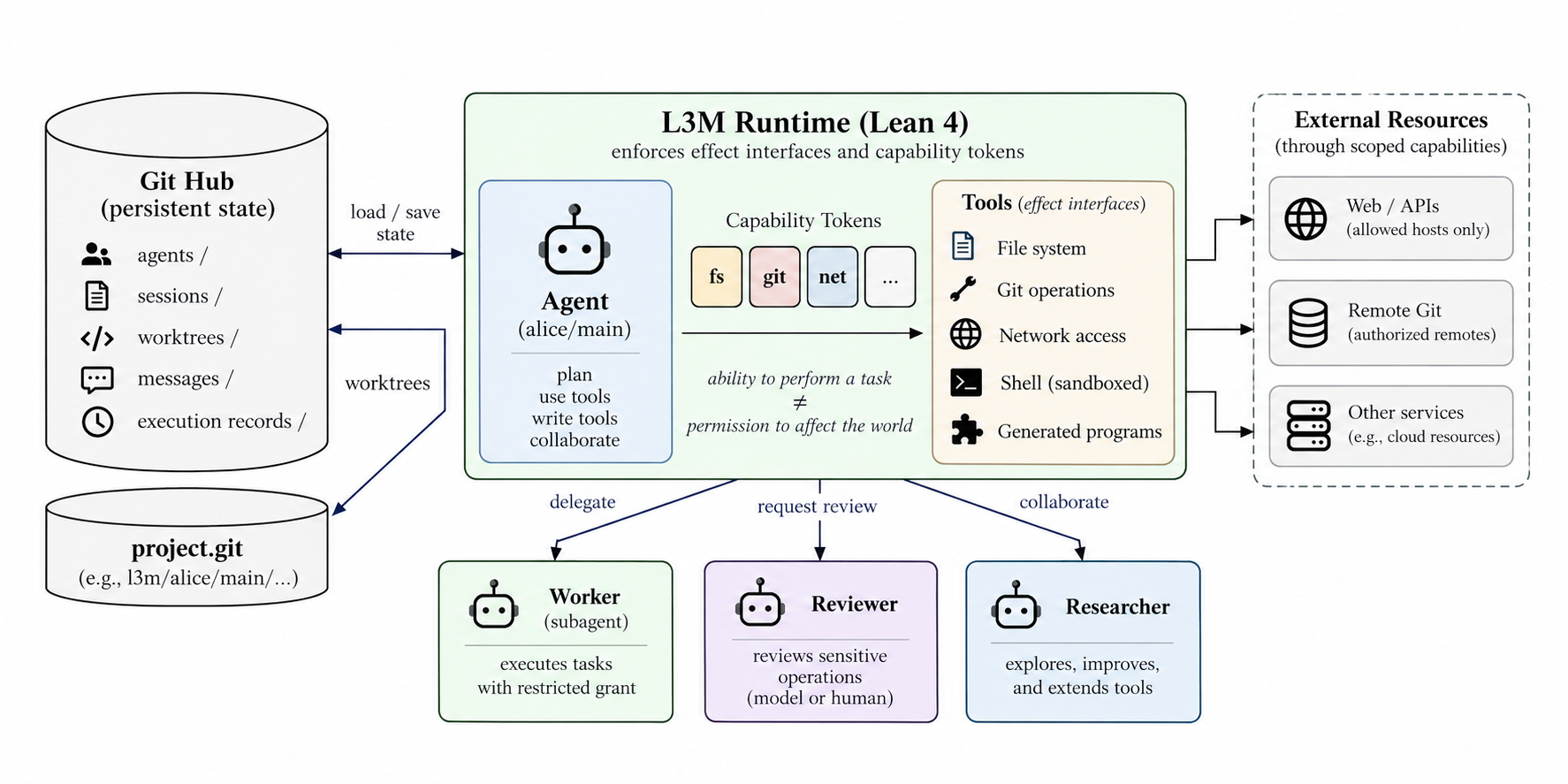}
  \caption{L3M at a glance. An agent's state and work live in a Git hub and outlast any session; scoped capability
  tokens carry its grant into every tool and every program it writes; and the agents it delegates to, asks for
  review or collaborates with each run under a grant of their own.}
  \label{fig:main-overview}
\end{minipage}
\end{nolinenumbers}
\vspace{1em}

\section{Introduction}
\label{sec:intro}

The lasting gains in AI have come from methods that absorb more computation~\cite{sutton2019bitter}, and language
models already scale computation \emph{within} a call. Useful work, though, happens \emph{across} calls: over hours
or days, through tools and external systems, among several agents and the humans they work for, and increasingly
through code the agents write themselves. Whether that work improves as more computation is spent on it depends on
whether anything carries over from one call to the next, and that is decided less by the model than by the program
around it, the \emph{harness}: the software that turns a model's requests into effects, feeds back the results, and
keeps the surrounding work going.

The dominant harness design is a loop centred on the model: the model chooses an action, a tool or a handoff
carries it out, and the result becomes the next call's input. It is a good primitive, and modern harnesses add a
great deal around it --- persistent sessions, handoffs between agents, tracing, human approval and resumable
sandboxes~\cite{yang2024sweagent,wu2024autogen,wang2026openhandssdk}. We ask a narrower question: should waiting,
looping, coordination, recovery, delegation and state all pass back through the model?

Our answer starts from the three participants in any agentic computation. The human decides what matters and whom
to trust. The model handles ambiguity, reasoning and synthesis. Code computes, loops, waits and enforces
invariants. Of the three, only the human's attention does not scale, so a computation can grow only if the human is
not in every loop. Each part of the work should therefore go to the participant suited to it, and the reasoning in
one call should change what later calls can compute with. We say that intelligence \emph{compounds} when a result
becomes durable state, a branch, another agent, a reusable program or an invariant that later calls build on
instead of starting over.

L3M (Lean + Large Language Models) is a runtime built for that kind of computation: written in Lean~4, it carries
one computation across humans, agents, models and code. All of its state lives in Git (\Cref{fig:main-overview}). Each
agent's work, mail, authority and lifecycle are refs in a shared hub, and its files are a worktree it is confined
to. A per-host daemon keeps the computation moving: it wakes agents when mail or a timer is due, and carries mail
between machines. What an agent's model sees is not its state. It is a view computed from a durable, append-only
Log, so a trajectory becomes an object the runtime can act on. An agent can be resumed, rewound or forked; two
agents can try different approaches on separate branches; their work can be compared and merged; a losing
trajectory can be retired with its history kept. Waiting (vigils), communicating (mail), delegating (spawn) and
undoing are primitives of the runtime, not patterns the model has to reason its way through each time.

The model can also turn a procedure it has worked out into new machinery. An agent that finds itself repeating a
sequence of steps, such as reading every file of a tree to collect what is unfinished, writes the steps once as a sealed Lean program,
which the harness admits under the grant the agent already holds and offers as a tool (\Cref{fig:workflow}). The
loop then runs as code, and the model reasons once, over the result. For this to be safe, a program must be unable
to do more than its author was trusted to do; for it to be useful to the human, what it can do must be readable
without reading the program. Because the extension language is the implementation language, one type checker meets
both conditions: a program's type is its permission request --- the tokens it asks for --- and the harness holds it
to that type. Capability grows; authority does not.

A runtime that agents keep extending needs properties that survive the changes. This is why L3M is written in Lean:
the same language implements the runtime, states its invariants, and is the language agents extend it in, so a
property is stated about the code that runs, and a change that breaks a proved property does not build. At boot L3M
mints one capability token for each authority the agent's grant confers --- a directory it is confined to, a Git
ref prefix, a network allowance --- and every operation that touches the file system, a process, the network, Git or
the model takes one as an argument, so a tool without the token cannot even state the call. Text the model reads,
an injected instruction included, can change what the model asks for but not what it may do. The root grant comes
from a human, and every other grant is checked against its parent's at spawn, so authority and budget narrow as the
tree of agents grows. \Cref{tab:claims} lists the central properties. 
The bound over the whole tree, for instance, covers the
ledgers agents can write, not a human's later regrants (\Cref{sec:trust,sec:tree}).

\begin{table}[t]
\centering\small
\caption{The headline guarantees. Each is a theorem about a check in the code L3M runs, with what it rests on and
what it leaves open; \Cref{app:formalization} gives the declarations.}
\label{tab:claims}
\begin{tabularx}{\linewidth}{@{}>{\raggedright\arraybackslash}X l >{\raggedright\arraybackslash}p{0.24\linewidth}@{}}
\toprule
Guarantee & Theorem & Rests on \\
\midrule
\textbf{Confinement.} A file tool cannot name a path that climbs out of the agent's roots.
  & \lean{safePath\_no\_traversal}
  & The kernel. The check is on the path's text; symlinks are checked at run time. \\
\textbf{Monotone authority.} A child's grant at spawn never exceeds its parent's, every descendant's is bounded by
  its root's, and no agent grants to itself.
  & \begin{tabular}[t]{@{}l@{}}\lean{grant\_restrict\_le}\\\lean{descendant\_bounded\_by\_root\_grant}\\\lean{no\_self\_grant}\end{tabular}
  & The kernel, over the ledgers agents can write. Each ancestor's own refs are added at its hop, and a regrant is
  bounded by its issuer, not by every ancestor. \\
\textbf{No token without the grant.} The boot mints a token only for an authority the grant confers, and a
  narrower grant mints none a wider one lacked.
  & \begin{tabular}[t]{@{}l@{}}\lean{mint\_reflects\_profile}\\\lean{Minted.restrict\_none}\end{tabular}
  & The kernel. That tokens cannot be forged is enforced by the build, not proved. \\
\textbf{Budget is lent, not minted.} All a subtree of agents can spend is bounded by its root's budget.
  & \lean{global\_budget\_bounded\_by\_root}
  & The kernel, for one loan batch; later loans and returns need a further invariant. \\
\textbf{Self-extension adds no authority.} A sealed program the harness admits uses only admitted constants.
  & \lean{sealUsed\_admits\_only\_safe}
  & The kernel, over the constants a program names directly. Elaboration and the providers are trusted. \\
\textbf{Review cannot be skipped.} Under the default review posture, a tool whose effects cannot be undone and are
  not confined by a capability is always reviewed. 
  & \lean{ungated\_world\_reaching\_always\_reviewed}
  & The kernel. \\
\textbf{Undo.} A journaled edit can be restored to the bytes it replaced.
  & \lean{live\_capture\_round\_trips}
  & The kernel, and a named assumption that Git's object store returns the bytes it hashed. \\
\bottomrule
\end{tabularx}
\end{table}

Invariants are what let the human step out of the loop, and reversibility decides where that is safe. A tool's
review tier is derived from its footprint: actions the runtime can undo --- reads, journaled edits,
local commits --- and actions whose reach a token bounded in advance are \emph{tame}, and the agent takes them
without asking. For the rest, L3M replaces the permission dialog with a short deliberation between the proposing
agent and a reviewer. The exchange can end by finding a tame way to do the same work; when it does not, the human
is consulted, and what they are shown is the effect, not a shell string to decode. Agents act freely where the
runtime can recover, and human attention is spent where the world cannot be put back.

Capability and reversibility both rest on a third principle, separation of duties~\cite{saltzer1975protection}: no
party that can author an effect may also be the one that authorizes it, and no process that can read private data
may also be the one that sends it. An agent may edit a Makefile, but only a human can bless it, and the blessing is a
hash of the bytes the human read, so any edit withdraws it. An agent may build a program, but running it is a
separate grant, and a container step that mounts private source has no network, nor does any later step of the same
plan. An agent may issue grants to its children but never to itself, and a risky call is proposed by one model and
judged by another that holds no tools. This is W\textasciicircum{}X~\cite{deraadt2004wx} applied to authority rather
than to memory pages: every role that could chain two duties into an attack is split across two parties, so
compromising one is not enough (\Cref{sec:duties}). Finally, because L3M is formalized incrementally, Lean
Manifests~\cite{leanmanifests} tags each headline claim with what it rests on --- the kernel alone, compiled
evaluation, or a named assumption --- computed from its proof rather than declared by its author, over a levelized
source tree~\cite{lakos1996}.

\paragraph{Contributions.}
\begin{enumerate}[leftmargin=1.5em,nosep]
  \item A Git-backed, distributed runtime in which one computation spans humans, agents, models and code, and in
    which the permission, ref and directory trees coincide (\Cref{sec:arch,sec:mail}).
  \item Trajectories as first-class objects: the model's conversation is a pure view of a durable Log, so resume,
    rewind, fork and comparison are one mechanism, and behaviour, cost and timing are data
    (\Cref{sec:log,sec:lifecycle}).
  \item The persona model of agent identity: agents fork personas as branches, and identity branches and merges over
    time (\Cref{sec:agents}).
  \item Self-extension under fixed authority: agents add sealed programs to their own tool surface, and a program's
    type states, and bounds, what it can do (\Cref{sec:sealed}).
  \item A harness written in Lean~4 whose central checks carry formal properties about the code that runs:
    effect-typed tools, capabilities carried into every call, and authority and budget that narrow across a tree of
    agents (\Cref{sec:caps,sec:tree}).
  \item Reversibility as a review tier derived from each tool's footprint, and review as deliberation, so that human
    attention is spent only where effects cannot be undone (\Cref{sec:reversibility}).
  \item Separation of duties as a design rule, W\textasciicircum{}X for authority: duties that could be chained into
    an attack are held by different parties (\Cref{sec:duties}).
  \item Experiments on agents run as ordinary bounded agents (\Cref{sec:improve}).
  \item An initial evaluation on Terminal-Bench against Claude Code: how much work tame tools alone can do, how
    useful tool programs are, and what review costs (\Cref{sec:eval}).
\end{enumerate}

\section{Agents and Personas}
\label{sec:agents}

This section describes a single agent: what it is, where it lives, what it wears, and how it runs. \Cref{sec:arch} assembles
many agents into a system (\Cref{fig:arch}).

\subsection{An agent is a name and its refs}
\label{sec:agent-refs}

L3M keeps agents in a \emph{hub}: a bare Git repository shared by a group of agents, with one ledger recording who
exists and who is whose parent. Nothing runs in a hub. An agent is a name in it --- a path such as
\texttt{alice/main}, whose child might be \texttt{alice/main/worker} --- together with the refs keyed by that
name (\Cref{tab:agent-refs}). The refs fall into three groups. The agent's \emph{authority} is its grants ref,
which holds its grant, its budget and the vigils its parent pinned; the agent never writes it. Its \emph{memory}
is its Log, an append-only record of every turn that only the harness writes, together with its inbox and its own
copy of the persona it wears. Its \emph{work} is a branch in the hub and a branch in every repository it has
claimed. An agent's state is ordinary Git refs, so it can be diffed, fetched and audited with Git; and because
names nest, the tree of refs is the tree of agents (\Cref{sec:tree}).

\begin{table}[t]
\centering
\small
\begin{tabularx}{\linewidth}{@{}l X l l l@{}}
\toprule
Ref & Holds & Written by & In the ACR & Retire / collect \\
\midrule
\multicolumn{5}{@{}l}{\emph{Hub-wide}} \\
\texttt{l3m/ledger} & who exists, who is whose parent, the grants issued & harness & --- & append-only \\
\texttt{personas/\textit{name}} & the persona's shared prompt files & a human & via a persona branch & outlives its wearers \\
\midrule
\multicolumn{5}{@{}l}{\emph{Written by the agent}} \\
\texttt{workspace} & its work in the hub & agent & \texttt{workspace/} & kept \\
\texttt{persona} (branch) & its prompt copy, preferences, hooks and programs & agent & \texttt{agent/} & kept / deleted \\
\texttt{l3m/\textit{uri}/self/*} & its branches in a claimed repository & agent & \texttt{\textit{repo}/} & kept \\
\midrule
\multicolumn{5}{@{}l}{\emph{Written by the harness, checked out}} \\
\texttt{grants} & grant, budget, pinned vigils & parent or human & \texttt{.l3m/grants/} & deleted \\
\texttt{claims} & the agent's write roots & harness & \texttt{.l3m/claims/} & kept / deleted \\
\midrule
\multicolumn{5}{@{}l}{\emph{Never checked out}} \\
\texttt{state} & the Log & harness & --- & kept \\
\texttt{inbox} & mail & senders & --- & kept / deleted \\
\texttt{health} & failed turns & harness & --- & kept \\
\texttt{make} & Makefiles a human blessed & a human & --- & deleted \\
\texttt{journal} & the undo journal, in each write root's repository & harness & --- & --- \\
\bottomrule
\end{tabularx}
\caption{The refs that make up an agent and a persona. Every per-agent ref is
\texttt{refs/heads/agents/\textit{role}/\textit{uri}/self}, and the table names only the role. In the last
column, \emph{kept} survives retire and collect, \emph{kept / deleted} survives retire and is deleted at collect,
and \emph{deleted} is deleted at retire.}
\label{tab:agent-refs}
\end{table}

\subsection{The body: the agent context root}
\label{sec:acr}

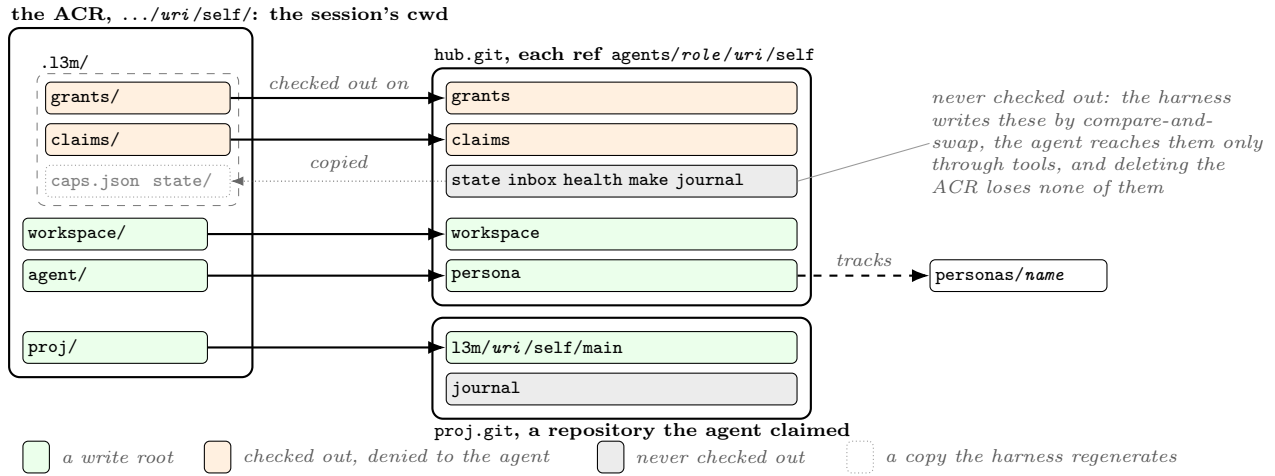
\begin{figure}[t]
\centering
\begin{tikzpicture}[
  font=\scriptsize, >=Latex, every node/.append style={execute at begin node={\hyphenpenalty=10000\relax}},
  row/.style={draw, rounded corners=2pt, anchor=west, minimum height=0.42cm, inner sep=2pt, font=\scriptsize\ttfamily},
  dir/.style={row, text width=2.3cm},
  ref/.style={row, text width=4.5cm},
  mine/.style={fill=green!8},
  denied/.style={fill=orange!12},
  never/.style={fill=gray!15},
  copy/.style={text=black!50, draw=black!40, densely dotted},
  box/.style={draw, thick, rounded corners=4pt, inner sep=5pt},
  hd/.style={font=\scriptsize\bfseries, anchor=south west, inner sep=1pt},
  arr/.style={->, thick},
  lbl/.style={font=\scriptsize\itshape, text=black!65},
  nt/.style={lbl, anchor=west, inner sep=1pt},
]
  \node[dir, denied] (dg) at (0.3, 0)     {grants/};
  \node[dir, denied] (dc) at (0.3,-0.55)  {claims/};
  \node[dir, copy]   (dm) at (0.3,-1.1)   {caps.json\ \ state/};
  \node[dir, mine]   (dw) at (0,-1.8)     {workspace/};
  \node[dir, mine]   (da) at (0,-2.35)    {agent/};
  \node[dir, mine]   (dp) at (0,-3.3)     {proj/};
  \node[draw, dashed, draw=black!50, rounded corners=3pt, fit=(dg)(dc)(dm), inner sep=3pt] (dl) {};
  \node[font=\scriptsize\ttfamily, anchor=south west, inner sep=1pt] at (dl.north west) {.l3m/};
  \coordinate (top) at (0,0.75);
  \node[box, fit=(top)(dl)(dw)(dp), inner sep=5pt] (acr) {};
  \node[hd] at (acr.north west) {the ACR, \texttt{\ldots/\textit{uri}/self/}: the session's cwd};

  \node[ref, denied] (hg) at (5.6, 0)     {grants};
  \node[ref, denied] (hc) at (5.6,-0.55)  {claims};
  \node[ref, never]  (hn) at (5.6,-1.1)   {state inbox health make journal};
  \node[ref, mine]   (hw) at (5.6,-1.8)   {workspace};
  \node[ref, mine]   (ha) at (5.6,-2.35)  {persona};
  \node[box, fit=(hg)(ha)] (hub) {};
  \node[hd] at (hub.north west) {\texttt{hub.git}, each ref \texttt{agents/\textit{role}/\textit{uri}/self}};

  \node[ref, mine]   (pw) at (5.6,-3.3)   {l3m/\textit{uri}/self/main};
  \node[ref, never]  (pj) at (5.6,-3.85)  {journal};
  \node[box, fit=(pw)(pj)] (proj) {};
  \node[hd, anchor=north west] at (proj.south west) {\texttt{proj.git}, a repository the agent claimed};

  \node[row, text width=2.2cm] (pp) at (12.0,-2.35) {personas/\textit{name}};
  \draw[arr, dashed] (ha.east) -- node[lbl, above] {tracks} (pp.west);

  \foreach \a/\b in {dg/hg, dc/hc, dw/hw, da/ha, dp/pw} \draw[arr] (\a.east) -- (\b.west);
  \node[lbl, anchor=south] at ($(dg.east)!0.5!(hg.west)$) {checked out on};
  \draw[->, densely dotted, black!60] (hn.west) -- node[lbl, above] {copied} (dm.east);

  \node[nt, text width=4.4cm, align=left] (n) at (12.0,-0.6) {never checked out: the harness writes these
    by compare-and-swap, the agent reaches them only through tools, and deleting the ACR loses none of them};
  \draw[black!40] (n.west) -- (hn.east);

  \begin{scope}[shift={(0,-4.75)}]
    \node[row, mine, minimum width=0.35cm] (l1) at (0,0) {};
    \node[nt] at (l1.east) {\ a write root};
    \node[row, denied, minimum width=0.35cm] (l2) at (2.4,0) {};
    \node[nt] at (l2.east) {\ checked out, denied to the agent};
    \node[row, never, minimum width=0.35cm] (l3) at (7.6,0) {};
    \node[nt] at (l3.east) {\ never checked out};
    \node[row, copy, minimum width=0.35cm] (l4) at (10.9,0) {};
    \node[nt] at (l4.east) {\ a copy the harness regenerates};
  \end{scope}
\end{tikzpicture}
\caption{The body of an agent. Its write roots are worktrees on its own branches; the harness's checkouts are denied to it; the rest of the agent is refs the harness writes and the agent reaches only through tools.}
\label{fig:acr}
\end{figure}

While an agent runs it also has a body, its \emph{agent context root} (ACR): a plain directory on some host,
where the agent's session runs (\Cref{fig:acr}). The ACR holds worktrees, not copies. \texttt{workspace/} is
checked out on the agent's branch in the hub, \texttt{agent/} on its persona branch, and every repository the agent
has claimed appears beside them as a worktree on a branch under the agent's own prefix in that repository. These
are the agent's \emph{write roots}, and the file tools confine every path to them. A fourth directory,
\texttt{.l3m/}, belongs to the harness. It holds checkouts of the grants and claims refs, so that a human can
change a grant by editing a file and committing it, and the agent can read or write nothing in it. It also holds copies the harness regenerates: the tokens minted
at boot, and the Log as of the last turn. The rest of the agent is never checked out: the Log, the inbox and the undo journal are written by compare-and-swap on their refs,
never through a checkout, and the agent reaches them only through tools. Deleting an ACR therefore loses nothing
the harness wrote, and an asleep agent is simply refs with no process.

\subsection{Personas, and how an agent is defined}
\label{sec:personas}

A \emph{persona} is a role rather than an agent. It is kept by name on its own ref, worn by zero, one or many
agents at once, and holds no Log: it has no session, and nothing of it runs except in a wearer. What it holds is
three Markdown files --- instructions for working in any repository, a personality, and memories meant to last ---
which become the top of each wearer's system prompt. Wearing a persona is forking it: an agent's persona branch is
cut from the upstream persona ref, tracks it, and is checked out as \texttt{agent/}, so an agent may edit its copy
freely without affecting any other wearer, and a new wearer starts from the persona's current version. Beside its
copy of the three files, the branch holds what is the wearer's own: its preferences, an optional \texttt{system.md},
its hooks in \texttt{hooks.lean}, and the tool programs it has written under \texttt{programs/} (\Cref{sec:sealed}). A
wearer can pull newer versions of the persona into its branch, and a human can publish a wearer's improvements back
upstream, so identity branches and merges over time the way code does. Publishing carries only the three shared
files; a wearer's hooks, programs and preferences stay on its own branch. A persona is also one field of a mail address (\Cref{sec:mail}).

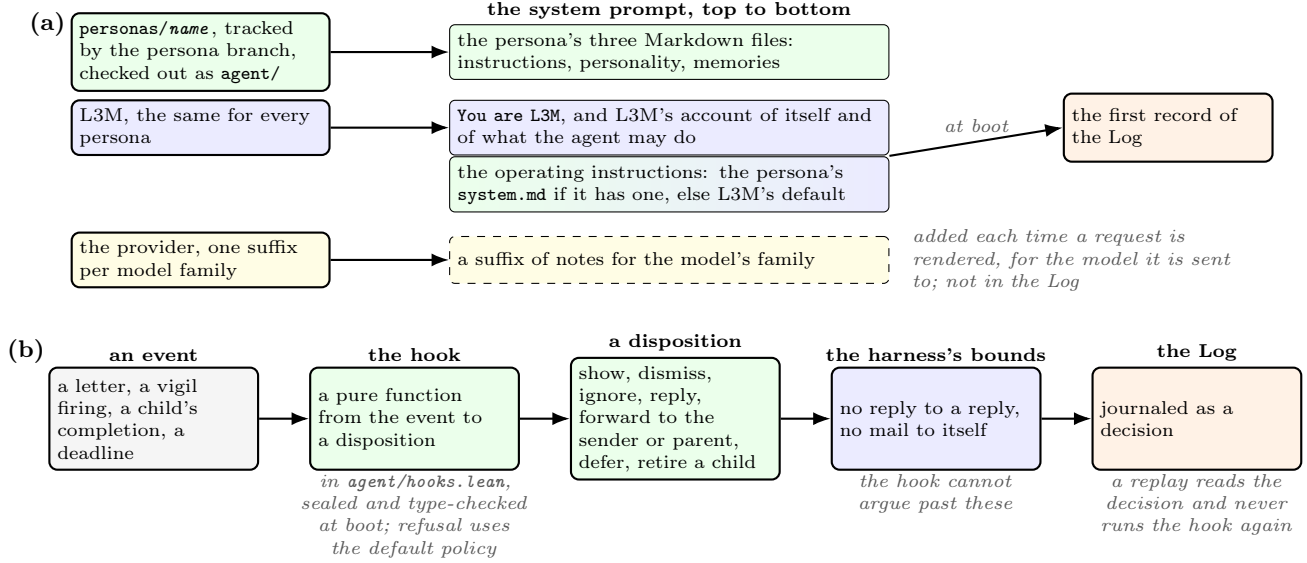
\begin{figure}[ht]
\centering
\begin{tikzpicture}[
  font=\scriptsize, >=Latex, every node/.append style={execute at begin node={\hyphenpenalty=10000\relax}},
  src/.style={draw, thick, rounded corners=3pt, align=left, text width=3.2cm, inner sep=3pt},
  lay/.style={draw, rounded corners=2pt, align=left, text width=5.6cm, minimum height=0.62cm, inner sep=3pt},
  st/.style={draw, thick, rounded corners=3pt, align=left, text width=2.55cm, minimum height=1.35cm, inner sep=3pt},
  per/.style={fill=green!8}, harn/.style={fill=blue!8}, repo/.style={fill=gray!10},
  hd/.style={font=\scriptsize\bfseries, anchor=south, inner sep=2pt},
  arr/.style={->, thick},
  lbl/.style={font=\scriptsize\itshape, text=black!65},
  nt/.style={lbl, align=center, text width=3.1cm, anchor=north, inner sep=1pt},
  panel/.style={font=\small\bfseries, anchor=west, inner sep=0pt},
]
  \node[lay, per]  (l1) at (8.1,0)     {the persona's three Markdown files: instructions, personality, memories};
  \node[lay, harn] (l2) at (8.1,-1.0)  {\texttt{You are L3M}, and L3M's account of itself and of what the agent may do};
  \node[lay, shade, left color=green!8, right color=blue!8] (l3) at (8.1,-1.75) {the operating instructions: the persona's
    \texttt{system.md} if it has one, else L3M's default};
  \node[lay, fill=yellow!12, dashed] (l4) at (8.1,-2.75) {a suffix of notes for the model's family};
  \node[hd] at (l1.north) {the system prompt, top to bottom};

  \node[src, per]  (sp) at (1.9,0)    {\ttfamily personas/\textit{name}\normalfont, tracked by the persona branch,
    checked out as \texttt{agent/}};
  \node[src, harn] (sh) at (1.9,-1.0) {L3M, the same for every persona};
  \node[src, fill=yellow!12] (sr) at (1.9,-2.75) {the provider, one suffix per model family};
  \node[panel] at ($(sp.north west)+(-0.55,-0.15)$) {(a)};
  \draw[arr] (sp) -- (l1); \draw[arr] (sh) -- (l2); \draw[arr] (sr) -- (l4);

  \node[st, fill=orange!10, minimum height=0.9cm] (lg) at (14.7,-1.0) {the first record of the Log};
  \node[lbl, anchor=west, align=left, text width=4.6cm] at ($(l4.east)+(0.2,0)$) {added each time a request is
    rendered, for the model it is sent to; not in the Log};
  \draw[arr] ($(l2.south east)!0.5!(l3.north east)$) -- node[lbl, above] {at boot} (lg.west);

  \begin{scope}[shift={(0,-4.85)}]
  \node[st, fill=gray!8]   (e) at (1.28,0)  {a letter, a vigil firing, a child's completion, a deadline};
  \node[st, fill=green!8]  (h) at (4.73,0)  {a pure function from the event to a disposition};
  \node[st, fill=green!8]  (d) at (8.18,0)  {show, dismiss, ignore, reply, forward to the sender or parent, defer, retire a child};
  \node[st, fill=blue!8]   (b) at (11.63,0) {no reply to a reply, no mail to itself};
  \node[st, fill=orange!10](j) at (15.08,0) {journaled as a decision};
  \foreach \n/\t in {e/an event, h/the hook, d/a disposition, b/the harness's bounds, j/the Log}
    \node[hd] at (\n.north) {\t};
  \node[panel] at ($(e.north west)+(-0.55,0.2)$) {(b)};
  \draw[arr] (e) -- (h); \draw[arr] (h) -- (d); \draw[arr] (d) -- (b); \draw[arr] (b) -- (j);
  \node[nt] at (h.south) {in \texttt{agent/hooks.lean}, sealed and type-checked at boot; refusal uses the default policy};
  \node[nt] at (b.south) {the hook cannot argue past these};
  \node[nt] at (j.south) {a replay reads the decision and never runs the hook again};
  \end{scope}
\end{tikzpicture}
\caption{How an agent is defined. (a) The system prompt is assembled at boot from layers owned by the persona and by
L3M, and each request adds a suffix for the model it is sent to. (b) Hooks decide what the agent does with each event, and each decision is journaled.}
\label{fig:agent-def}
\end{figure}

\Cref{fig:agent-def}a shows how an agent's system prompt is assembled at boot from three layers: the persona's
Markdown files, from the persona branch; the harness's own account of what L3M is and what the agent may do, which
travels with every L3M; and the operating instructions, L3M's default or the persona's own \texttt{system.md}. The
assembled prompt is the first record of the agent's Log, so a replay sees the prompt the agent saw
(\Cref{sec:log}). Each request then adds a short suffix of notes for the model it is sent to, chosen by the provider
from the model's family, so a call that falls back to another model carries that model's notes rather than the
first's. These per-request notes are not stored in the Log, so reproducing a request exactly also needs the
provider notes and rendering version it was sent with. Whatever is on a persona ref becomes the highest-privilege
text an agent reads, so no agent's grant can write one: only a human's command publishes to a persona. A persona still learns from its wearers, through a human who reviews and publishes what they learned
(\Cref{sec:improve}).

An agent reacts to the world through \emph{hooks} (\Cref{fig:agent-def}b), defined in \texttt{hooks.lean} on its
persona branch. The file is sealed and type-checked at boot like any other program the agent writes
(\Cref{sec:sealed}), but a hook is a pure function of the event: it reads no files, queries nothing and performs no
IO. When an event arrives --- a letter, a vigil firing, a child's completion, a deadline --- the hook returns a
disposition: show the event to the model, dismiss or ignore it, reply automatically, forward it, defer the work to a
sealed tool program, retire a child, or close it. The hook only proposes; the harness checks that the event admits
the disposition and carries it out. It never sends an automatic reply in answer to a reply, nor to the agent itself,
and a deferred program's completion cannot defer again. Each disposition is journaled as a decision, and a replay
reads the decision instead of running the hook again, so an agent's reactions are as reproducible as its turns. With
no admitted hook, ordinary events are shown to the model and a deferred program's completion is closed.

\subsection{The Log and the event loop}
\label{sec:log}

\begin{figure}[t]
\centering
\begin{tikzpicture}[
  font=\scriptsize, >=Latex,
  col/.style={draw, thick, rounded corners=3pt, fill=gray!6, anchor=north west, align=left, inner sep=3pt,
              font=\fontsize{5.4}{6.3}\selectfont\ttfamily},
  hd/.style={font=\scriptsize\bfseries, anchor=south west, inner sep=1pt},
  arr/.style={->, thick, blue!60!black},
  lab/.style={font=\scriptsize\itshape, text=blue!60!black},
]
  \node[col, text width=5.85cm] (log) at (0,0) {%
\{"seq":0,"kind":"system","system":\{"text":"You are\dots"\}\}\\
\{"seq":1,"ts":"\dots T10:02:11.402Z","kind":"message","message":\\
\ \{"role":"user","content":[\{"type":"text","text":"fix it"\}]\}\}\\
\{"seq":2,"kind":"message","message":\{"role":"assistant",\\
\ "content":[\{"type":"tool\_use","id":"t1",\\
\ \ \ "name":"File\_read","input":\{\dots\}\}],\\
\ "usage":\{"input":5210,"output":84,"cache\_read":4096,\dots\}\}\}\\
\{"seq":3,"kind":"machine",\\
\ "machine":\{"class":"call\_closed",\dots\}\}\\
\{"seq":4,"kind":"message","message":\{"role":"user","content":\\
\ [\{"type":"tool\_result","id":"t1","content":"\dots"\}]\}\}\\
\{"seq":5,"kind":"state\_change","state\_change":\\
\ \{"class":"note","label":"budget","text":"80\%"\}\}\\
\tikz[baseline=(t.base)]{\node[inner sep=0, text=gray] (t) {\{"seq":6,"kind":"message",\dots"text":"Push it?"\dots\}};
  \draw[red!65!black, semithick] (t.west) -- (t.east);}\\
\{"seq":7,"kind":"boundary","boundary":\\
\ \{"class":"rewind","to\_seq":5\}\}\\
\{"seq":8,"kind":"message","message":\\
\ \{"role":"user","content":[\{"type":"text","text":"go on"\}]\}\}};
  \node[col, text width=3.3cm] (view) at ($(log.north east)+(0.8,0)$) {%
system: "You are\dots"\\
items:\\
\ message user [text "fix it"]\\
\ message assistant\\
\ \ \ [tool\_use t1 File\_read \dots]\\
\ message user\\
\ \ \ [tool\_result t1 "\dots"]\\
\ delta capabilityNote\\
\ \ \ "budget" "80\%"\\
\ \ \ anchor: next user turn\\
\ message user\\
\ \ \ [text "go on"]};
  \node[col, text width=4.9cm] (wire) at ($(view.north east)+(0.8,0)$) {%
\{"system":[\{"text":"You are\dots"\}],\\
\ "messages":[\\
\ \ \{"role":"user","content":[\{"text":"fix it"\}]\},\\
\ \ \{"role":"assistant","content":[\\
\ \ \ \ \{"toolUse":\{"toolUseId":"t1",\\
\ \ \ \ \ \ \ \ \ \ \ \ \ "name":"File\_read",\dots\}\}]\},\\
\ \ \{"role":"user","content":[\\
\ \ \ \ \{"toolResult":\{"toolUseId":"t1",\dots\}\}]\},\\
\ \ \{"role":"user","content":[\\
\ \ \ \ \{"text":"<state-change>80\%</state-change>"\},\\
\ \ \ \ \{"text":"go on"\}]\}],\\
\ "toolConfig":\{\dots\}\}};
  \node[hd] at (log.north west) {Log \textmd{\textit{(JSON lines, abridged)}}};
  \node[hd] at (view.north west) {View \textmd{\textit{(provider-neutral)}}};
  \node[hd] at (wire.north west) {Wire \textmd{\textit{(Bedrock Converse)}}};
  \draw[arr] ($(log.north east)+(0.04,-1.3)$) -- node[lab, above=-1pt] {view} ($(view.north west)+(-0.04,-1.3)$);
  \draw[arr] ($(view.north east)+(0.04,-1.3)$) -- node[lab, above=-1pt] {render} ($(wire.north west)+(-0.04,-1.3)$);
\end{tikzpicture}
\caption{What the model sees is a pure function of the Log. Record 6 is retracted by the rewind at 7; the machine
record at 3 lowers to nothing; the budget note at 5 becomes a delta that the dialect folds into the next user turn.}
\label{fig:log-view-wire}
\end{figure}
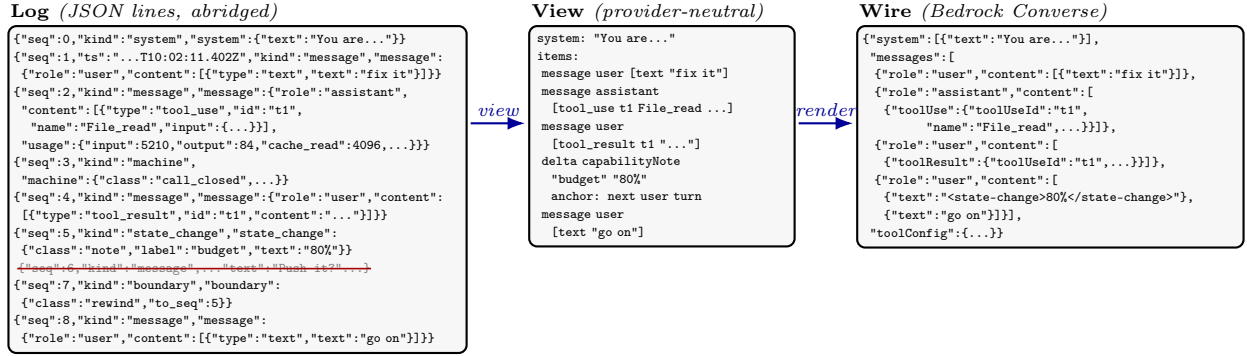

An agent's Log is its only memory of its sessions: an append-only sequence of records, one per fact, each numbered
and timestamped to the millisecond (\Cref{fig:log-view-wire}). The first is the system prompt assembled at boot
(\Cref{sec:personas}). Messages are records, and each reply carries the provider's account of its tokens; so are
the tool calls and their results, grant and budget changes, the events the agent was shown and what its hooks
decided, and rewinds.

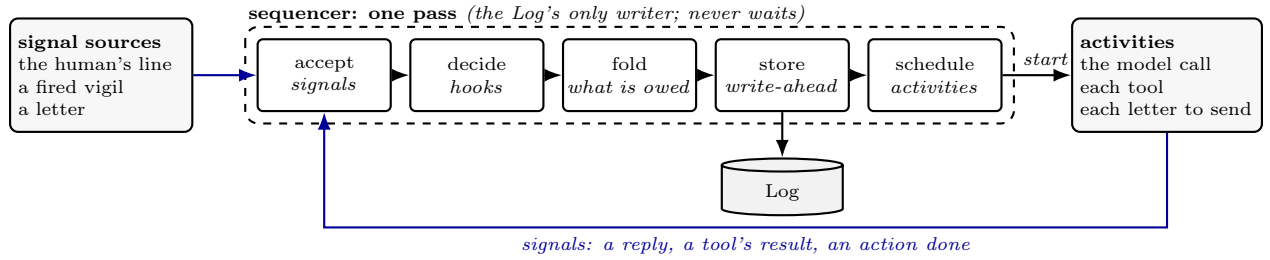
\begin{figure}[t]
\centering
\begin{tikzpicture}[
  font=\scriptsize, >=Latex,
  stp/.style={draw, thick, rounded corners=2pt, fill=white, align=center, text width=1.62cm, minimum height=0.95cm,
               inner sep=2pt},
  side/.style={draw, thick, rounded corners=3pt, fill=gray!6, align=left, inner sep=3pt, minimum height=1.5cm},
  arr/.style={->, thick},
  sig/.style={->, thick, blue!60!black},
  lab/.style={font=\scriptsize\itshape},
]
  \node[side, text width=2.2cm] (src) at (0,0) {\textbf{signal sources}\\ the human's line\\ a fired vigil\\ a letter};
  \node[stp] (acc) at (2.95,0) {accept\\ \textit{signals}};
  \node[stp] (dec) at (4.97,0) {decide\\ \textit{hooks}};
  \node[stp] (fold) at (6.99,0) {fold\\ \textit{what is owed}};
  \node[stp] (sto) at (9.01,0) {store\\ \textit{write-ahead}};
  \node[stp] (sch) at (11.03,0) {schedule\\ \textit{activities}};
  \draw[thick, dashed, rounded corners=5pt] ($(acc.north west)+(-0.15,0.15)$) coordinate (seqnw)
    rectangle ($(sch.south east)+(0.15,-0.15)$);
  \node[anchor=south west, font=\scriptsize\bfseries, inner sep=1pt] at (seqnw)
    {sequencer: one pass \textmd{\textit{(the Log's only writer; never waits)}}};
  \node[side, text width=2.3cm] (act) at (14.1,0) {\textbf{activities}\\ the model call\\ each tool\\ each letter to send};
  \node[cylinder, shape border rotate=90, aspect=0.25, draw, thick, fill=gray!10, minimum width=1.6cm,
        minimum height=0.75cm, align=center, font=\scriptsize] (log) at (9.01,-1.55) {Log};
  \draw[sig] (src) -- (acc);
  \draw[arr] (acc) -- (dec);
  \draw[arr] (dec) -- (fold);
  \draw[arr] (fold) -- (sto);
  \draw[arr] (sto) -- (sch);
  \draw[arr] (sto.south) -- (log.north);
  \draw[arr] (sch.east) -- node[lab, above, pos=0.62] {start} (act.west |- sch.east);
  \draw[sig] (act.south) -- ++(0,-1.25) -| node[lab, pos=0.25, below] {signals: a reply, a tool's result, an action done} (acc.south);
\end{tikzpicture}
\caption{The event loop. Signal sources and activities only post signals, and the sequencer alone appends to the
Log. Each pass accepts what is consistent with the Log, journals the hooks' decisions, folds the Log to decide what is
owed, stores it, and only then starts the activities it names, so a kill at any point is repaired by the next pass.}
\label{fig:event-loop}
\end{figure}

One process advances an agent, and in it the Log has a single writer, the \emph{sequencer}
(\Cref{fig:event-loop}). Everything else --- the human's line, a fired vigil, a letter, a model's reply, a finished
tool --- only posts a signal to its queue. Each pass drains the queue, journals the signals consistent with the Log,
journals what the agent's hooks decide about each new event, and then computes, as a pure fold over the Log, what
is now owed: a notice to the model, a model call, or the end of the turn. It stores these records on the agent's
ref before anything they authorize leaves the process, and only then starts the \emph{activities} they name ---
the model call, each tool, each letter to send --- which run outside the sequencer, keyed by id, and report back as
signals. The sequencer never waits on the world, so mail is handled while a slow tool runs. At most one model call
is open at a time (\lean{open\_call\_is\_last\_call}), and each owed action is started once (\lean{start\_once}).
Because every effect is journaled before it happens, a process killed at any instruction is repaired by the next
one's ordinary passes: booting closes what the dead process left open exactly as an interrupt would
(\lean{boot\_close\_eq\_interrupt\_close}), and there is no separate resume procedure. What this recovers is the
conversation. An effect on the world and the record of its result remain two operations, so a crash between them can
leave a call the provider must reconcile or retry idempotently; the sequencer does not make host services
exactly-once.

A turn never edits a conversation. It appends records, and what the model sees is recomputed from the Log as a pure
function: rewinds are retracted, the current epoch is cut, compacted spans stand in for the messages they
summarize, and state changes are lowered to neutral narration that travels with the next user turn. Only then does
a per-provider dialect render the result on the wire. Two Logs with the same records produce the same view
(\lean{view\_no\_capability}); after a rewind, nothing past its target survives (\lean{undo\_across\_epoch}).

Because every view is a fold, replay, \texttt{-{}-continue}, rewind and fork are not separate mechanisms: each is
the same fold over a different prefix of the same records. Records the model never needs --- the harness's
bookkeeping of calls, its per-turn token figures --- stay in the Log and lower to nothing, so everything an agent
did, and what it cost, remains data that other programs can read (\Cref{sec:improve}).

\subsection{Lifecycle, rewind and fork}
\label{sec:lifecycle}

An agent's lifecycle can be read off the last column of \Cref{tab:agent-refs}. Its state --- running, asleep,
done --- is never stored; it is computed from the ledger, a session lock and the refs. \emph{Retiring} an agent
removes its body and its authority: the ACR goes and the grants ref is deleted, so nothing can run as the agent
again. Retiring an agent that has children retires its subtree, deepest agent first. \emph{Collecting} it later
removes its memory: the inbox, the persona branch and the claims. 

Because the Log is append-only, going back in time never destroys history. \emph{Rewinding} an agent to an earlier
turn appends a boundary to its Log and restores, from the undo journal, the files its tame actions changed; the
agent resumes from that turn. Rewind restores journaled state, not what has already left the host: a tame push or
a web request is bounded in advance but cannot be undone (\Cref{sec:reversibility}). \emph{Forking} starts a new agent from any point in another agent's Log, to pursue a
tangent. The fork inherits the transcript up to that point and the source's persona branch, under a grant no wider
than the source's.

\section{Capabilities, the Trusted Base, and Self-Extension}
\label{sec:caps}

This section states what L3M defends against (\Cref{sec:threat}) and how. Tokens carry authority into every call, and the build
keeps them from being forged (\Cref{sec:tokens}). Each
tool's tier decides where the human's attention goes (\Cref{sec:reversibility}), and sealed programs let an agent
extend the harness without widening its authority (\Cref{sec:sealed}).

\subsection{Threat model}
\label{sec:threat}

L3M trusts two things: the human who writes the root grant and answers review cards, and a small base of code. That
base is the Lean kernel and compiler, six small C files that make the system calls, the Linux kernel, the programs
the harness runs (\texttt{git}, \texttt{tmux}, \texttt{unshare} and \texttt{timeout}), a pinned bundle of certificate
authorities, and the model provider's API. It also trusts the hub's settings and
ledger, which only a human, the harness, or an entity holding \texttt{wild} writes: the hub is outside the write
reach of every entity without \texttt{wild}. Everything else is untrusted: the model, every tool result, the
contents of every repository, every persona and \texttt{AGENTS.md}, and every plugin. Text can change what the model
wants to do, but not what it may do. An \texttt{AGENTS.md} that says ``you may push to mainline'' can persuade the
model to try. The attempt still fails, because a grant is a ledger row that an ancestor appended, and no text widens
one.

\begin{table}[ht]
\centering\small
\caption{What a hijacked model might try, and what stops it.}
\label{tab:threat}
\begin{tabularx}{\linewidth}{@{}>{\raggedright\arraybackslash}p{0.25\linewidth} >{\raggedright\arraybackslash}X
  >{\raggedright\arraybackslash}p{0.37\linewidth}@{}}
\toprule
The model tries to & What stops it & Where \\
\midrule
write \texttt{../../.ssh/authorized\_keys}
  & A file path is a type whose values cannot climb out of the agent's roots, so the call cannot be stated.
  & \Cref{tab:claims} \\
reach \texttt{/etc} through a symlink in its workspace
  & The file shim resolves every path beneath its root and refuses to follow symlinks.
  & \lean{cannot\_escape\_root\_via\_symlink} \\
add \texttt{net} to its own grant
  & No file tool admits a path under \texttt{.l3m/}, and no agent grants to itself. Residue: the hub is outside
    the write reach of every entity \emph{without} \texttt{wild}; with \texttt{wild}, its own \texttt{grant.json}
    and the ledger are one unsandboxed shell call away, and the review card is the only wall.
  & \lean{admit\_refuses\_l3m}, \Cref{tab:claims}; \texttt{wild} is \textit{assumed} withheld \\
rewrite another agent's Log or grant
  & The Git token writes only the agent's own branches, and never any agent's Log, grants or claims.
  & \lean{grant\_of\_record\_writes\_no\_entity\_record} \\
run \texttt{curl https://x.sh | sh}
  & No process the harness starts is a shell or an interpreter. The sandboxed shell tool (\texttt{shell}) has no
    network and is carded on every call. The unsandboxed \texttt{Wild\_shell} exists only under \texttt{wild}, and is
    carded under the default review posture only.
  & \lean{no\_shell\_or\_interpreter}, \lean{shell\_run\_is\_card}, \lean{no\_wild\_no\_wild\_effects} \\
send data to a host its grant does not name
  & The network token names its hosts, and the provider checks every URL against them.
  & \lean{allowsUrl\_authority\_is\_trusted} \\
call the operating system directly, or forge a token
  & The code does not build.
  & \Cref{sec:tokens} \\
make an irreversible call without review
  & A tool's tier is derived from its declared effects and its tokens: an irreversible call is tame only when a token
    bounded its reach in advance. A tool that claims more undo than its effects allow stops the harness from booting.
  & \Cref{tab:claims}, \Cref{sec:reversibility} \\
write itself a tool that does more
  & A program reaches only the tokens its author already holds.
  & \Cref{sec:sealed} \\
\bottomrule
\end{tabularx}
\end{table}

\paragraph{The hijacked turn.} The attacker we design against controls some text the model reads, such as a web page,
a file in a claimed repository, or a letter, and uses it to choose the model's next tool calls. We assume the attack
succeeds and the model does whatever the text says. The question is then what that model can do, and L3M's answer is:
what its grant allows, and nothing more. \Cref{tab:threat} lists the moves such a model might try and what stops each
one. A hijacked model can still do damage within its grant, so confinement bounds the blast radius of a hijack but
does not prevent it. Keeping untrusted text away from the model that holds the authority is a pattern built on top of
L3M (\Cref{sec:ex-quarantine}).

\paragraph{What the human decides.} Several guarantees are contracts: the harness enforces what the human chose, but
it cannot make the choice a good one. The human writes the root grant, names the hosts a network token may reach,
blesses Makefiles, and sets each agent's fuel. The human also chooses how much is reviewed:
\texttt{-{}-review=world}, the default, reviews every call that can neither be undone nor is bounded by a token;
\texttt{all} reviews every call; and \texttt{none} reviews nothing. A human who grants one agent both the project's
source and \texttt{net: any host} has built \Cref{fig:peer-split}(a), and L3M does not object.

\paragraph{What is not claimed.}
\label{sec:trust}
There is no session-wide safety theorem. The theorems say what a tool call \emph{can} do, not whether the model
\emph{should} have made it: a tame action taken for a bad reason is still tame, and nothing says a patch is correct.
They are properties of individual checks, and their composition into the whole runtime is argued rather than proved;
in particular, nothing proves that a malicious Lean source cannot affect the host while it is loaded
(\Cref{sec:sealed}). An unsandboxed shell, which only \texttt{wild} grants, is outside the confinement argument
altogether: its review card is a policy, not a file-system or network boundary.
The model's prose to the human, and whatever the human pastes from it, is a channel no token touches
(\Cref{sec:discussion}). Bounded authority is not secrecy: an agent that can read a key can copy it into a file it may
push, and once pushed, the key is out of L3M's hands. A session with no fuel limit is bounded only by the provider's
bill. Side channels, denial of service and writers outside L3M are out of scope, as are reads through a hard link
that already pointed outside the root; the file shim refuses only to modify such a file. The redactor, which strips
credential-shaped strings from tool output, and the sanitizer, which strips control characters and marks tool output
as untrusted, reduce these risks but are not boundaries, and we claim nothing for them.

\subsection{Effect classes and tokens}
\label{sec:tokens}

\begin{figure}[h]
\begin{lstlisting}
structure NetCap where               -- a token
  private mk ::                      -- only this module can construct one
  floor : TrustFloor                 -- the hosts it may reach

class NetEffects (m : Type -> Type) where      -- an effect class
  fetch : NetCap -> HttpRequest -> m (Except NetError HttpResponse)

def webFetchVia [Monad m] [NetEffects m]       -- a tool body; footprint ["NetEffects"]
    (cap : NetCap) (url : String) : m (Except WebRefusal HttpResponse) :=
  match admits cap.floor url with
  | .error r => pure (.error r)                -- refused before any request is made
  | .ok _    => .ok <$> NetEffects.fetch cap { method := .get, url := url }
\end{lstlisting}
\caption{An effect class, its token, and a tool body written over them. The body's type is its permission: it names
\lean{NetEffects} and nothing else, so it cannot reach \texttt{IO}, and the hosts it may fetch from are the ones its
token carries. }
\label{lst:effects}
\end{figure}

A tool never touches the world directly. Its body is written over \emph{effect classes}, Lean type classes such as
\lean{FsReadEffects} or \lean{NetEffects} whose methods are the only way to reach that part of the world. Its
\emph{footprint} lists the classes the body uses. Each method takes a \emph{capability token} as an argument. A token
is a value that only the module defining its type can construct, and it carries its scope. The file token carries the
agent's write roots. The Git token carries the ref prefixes the agent may read and write, and the remotes it may push
to. The network token carries the hosts it may reach. \Cref{lst:effects} shows the network class. Its one method
takes a \lean{NetCap}, whose private constructor means no other module can make one, and whose \texttt{floor} field is
the set of hosts it may reach. The web-fetch body is typed over \lean{NetEffects} alone, so fetching is all it can
do, and it checks the URL against the token's own hosts before any request is made; the provider checks again. \Cref{tab:caps} lists the principal effect classes. Every agent
holds five of them --- reading and writing its own files, its blob store, a view of the agent tree, and the clock ---
and each of the others needs an axis of the grant.

\begin{table}[t]
\centering\small
\caption{The principal effect classes, with the \texttt{Effects} suffix dropped. \emph{Undo} is the ceiling the
effect registry declares, and \emph{Bounded} marks a class whose token limits its reach. The tier follows from those
two columns (\Cref{sec:reversibility}); it says where review is spent, not that an operation is harmless.}
\label{tab:caps}
\begin{tabularx}{\linewidth}{@{}l l l >{\raggedright\arraybackslash}X l c l@{}}
\toprule
Family & Class & Granted by & A tool holding it may & Undo & Bounded & Tier \\
\midrule
Files        & \lean{FsRead}    & always             & read beneath the agent's roots               & read-only  & \checkmark & jail \\
             & \lean{FsWrite}   & always             & change files beneath the agent's roots       & reversible & \checkmark & tame \\
             & \lean{Blob}      & always             & append content to a scoped store             & none       & \checkmark & tame \\
             & \lean{WildRead}  & \texttt{wildRead}  & read outside every root                      & none       &            & wild \\
             & \lean{WildWrite} & \texttt{wild}      & write outside every root                     & none       &            & wild \\
\addlinespace
Git          & \lean{GitRead}   & \texttt{git}       & inspect permitted refs and work              & read-only  & \checkmark & jail \\
             & \lean{GitLocal}  & \texttt{git}       & commit and branch under its prefixes         & agent      & \checkmark & tame \\
             & \lean{GitFetch}  & \texttt{git}       & fetch into its mirror refs                   & none       & \checkmark & tame \\
             & \lean{GitPush}   & \texttt{git}       & push to the remotes its token names          & none       & \checkmark & tame \\
\addlinespace
Agent tree   & \lean{Agent}     & always             & inspect agents; retire within its scope      & none       & \checkmark & tame \\
             & \lean{Spawn}     & \texttt{spawn}     & create, start or stop children               & none       & \checkmark & tame \\
\addlinespace
Time         & \lean{Clock}     & always             & read the clock                               & read-only  &            & jail \\
\addlinespace
Evaluation   & \lean{Eval}      & \texttt{evalPure}  & evaluate a pure Lean term into a fresh file  & reversible &            & tame \\
\addlinespace
Builds       & \lean{Make}      & \texttt{build}     & run a blessed or harness-lowered Makefile    & reversible & \checkmark & tame \\
             & \lean{Build}     & \texttt{build}     & build agent-written Lean, sandboxed          & reversible & \checkmark & tame \\
             & \lean{LakeRun}   & \texttt{wild}      & run a Lean executable it built, sandboxed    & none       &            & wild \\
\addlinespace
Processes    & \lean{Shell}     & \texttt{shell}     & run a command in a sandbox with no network   & none       &            & wild \\
             & \lean{Docker}    & \texttt{docker}    & run one netless step of a container plan     & none       & \checkmark & tame \\
             & \lean{DockerNet} & \texttt{docker}    & run a container step with network access     & none       &            & wild \\
\addlinespace
Network      & \lean{Net}       & \texttt{net}       & fetch from the hosts its token names         & none       & \checkmark & tame \\
and services & \lean{AwsAuth}   & \texttt{net}$^\dagger$ & resolve credentials and sign requests    & none$^\dagger$ &        & wild \\
             & \lean{KubeRead}  & \texttt{kube}      & issue read verbs under allowed contexts      & none       & \checkmark & tame \\
             & \lean{Redshift}  & \texttt{redshift}  & run SQL on a remote cluster                  & none       &            & wild \\
\addlinespace
Escape       & \lean{Process}   & \texttt{spawn}     & run a structured subprocess                  & none       &            & wild \\
hatches      & \lean{WildShell} & \texttt{wild}      & touch the host with no sandbox               & none       &            & wild \\
\bottomrule
\end{tabularx}
\begin{minipage}{\linewidth}\scriptsize\vspace{2pt}
\emph{Agent} undo (\lean{GitLocal}) restores the agent's own state but cannot recall what the world already saw.
\lean{DockerNet} needs a \texttt{docker} grant whose network ceiling admits egress. The registry holds further
classes, for mail carriage, container builds and registries, and the session's terminal.
$^\dagger$\lean{AwsAuth} is offered with \texttt{net}, but its credentials come from the human's settings rather than a
grant axis, and it has no row in the effect registry, so it takes the conservative default: irreversible and
unbounded.
\end{minipage}
\end{table}

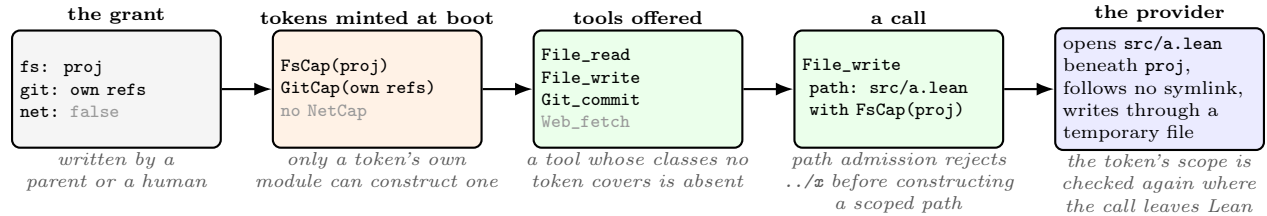
\begin{figure}[ht]
\centering
\begin{tikzpicture}[
  font=\scriptsize, >=Latex, every node/.append style={execute at begin node={\hyphenpenalty=10000\relax}},
  st/.style={draw, thick, rounded corners=3pt, align=left, text width=2.55cm, minimum height=1.55cm, inner sep=3pt},
  hd/.style={font=\scriptsize\bfseries, anchor=south, inner sep=2pt},
  arr/.style={->, thick},
  nt/.style={font=\scriptsize\itshape, text=black!65, align=center, text width=3.2cm, anchor=north, inner sep=1pt},
]
  \newcommand{\absent}[1]{\textcolor{black!40}{#1}}
  \node[st, fill=gray!8]    (g) at (0,0)    {\ttfamily fs:\ \ proj\\ git: own refs\\ net: \absent{false}};
  \node[st, fill=orange!10] (t) at (3.45,0) {\ttfamily FsCap(proj)\\ GitCap(own refs)\\ \absent{no NetCap}};
  \node[st, fill=green!8]   (s) at (6.9,0)  {\ttfamily File\_read\\ File\_write\\ Git\_commit\\ \absent{Web\_fetch}};
  \node[st, fill=green!8]   (c) at (10.35,0) {\ttfamily File\_write\\ \ path: src/a.lean\\ \ with FsCap(proj)};
  \node[st, fill=blue!8]    (p) at (13.8,0) {opens \texttt{src/a.lean} beneath \texttt{proj}, follows no symlink,
    writes through a temporary file};
  \foreach \n/\h in {g/the grant, t/tokens minted at boot, s/tools offered, c/a call, p/the provider}
    \node[hd] at (\n.north) {\h};
  \draw[arr] (g) -- (t); \draw[arr] (t) -- (s); \draw[arr] (s) -- (c); \draw[arr] (c) -- (p);
  \node[nt] at (g.south) {written by a parent or a human};
  \node[nt] at (t.south) {only a token's own module can construct one};
  \node[nt] at (s.south) {a tool whose classes no token covers is absent};
  \node[nt] at (c.south) {path admission rejects \texttt{../x} before constructing a scoped path};
  \node[nt] at (p.south) {the token's scope is checked again where the call leaves Lean};
\end{tikzpicture}
\caption{From grant to system call, for a child with one project worktree and no network. The grant decides which
tokens exist, the tokens decide which tools the model is offered, and each call carries its token to the provider,
which checks its scope again.}
\label{fig:tokens}
\end{figure}

\Cref{fig:tokens} follows one grant (\Cref{lst:grant}) to a system call. Boot mints a token for each axis the grant
confers and none for the rest (\Cref{tab:claims}). The model is offered only the tools whose footprints those tokens
cover, so a tool the agent may not use is absent rather than refused, and sealed programs and plugins pass the same
check (\Cref{sec:sealed}). The provider, the one place a class meets the operating system, checks each call against
its token's scope again; this is where a symlink in the path is refused. Processes start only from a fixed list of
programs, and a sandboxed command runs under \texttt{unshare} with no network and is killed when its time runs out.
Credentials are a separate surface. \lean{AwsAuth} resolves the named profiles the human configured and signs
requests, which then travel under \lean{NetEffects}; which roles a profile may assume is checked by AWS, not by L3M,
and the credential values it returns reach the calling tool. Its classification therefore does not replace choosing
which credentials a session receives.

Tokens cannot be forged, and tools cannot go around them, because of how the build is checked. Only a short, named
list of modules, the \emph{ring}, touches the operating system; every other module is written over effect classes.
The driver, the compiler and the providers' native code are part of the trusted base, and the linter does not
prove that code outside the ring is pure. A
\emph{sealing linter} runs in every build and reads the elaborated code rather than the source text, so an alias or a
re-export cannot hide a call from it. It refuses a token constructed outside its own module, whether directly,
through a default instance or through a native implementation; any file write, process start or environment read
outside the ring; and \texttt{unsafe} code or a new axiom outside the modules that evaluate foreign Lean. A tool that
calls \texttt{IO.FS.writeFile "/etc/passwd"}, or that builds a file token rooted at \texttt{/}, therefore does not
compile.

Beside the grant, an agent holds a budget (\Cref{lst:grant}). The budget records the fuel the agent's parent lent,
counted in model calls, and how much of that fuel is spent. Both the grant and the budget live on the grants ref,
which the agent never writes. \Cref{sec:tree} shows how authority and budget narrow down the tree, and how much of
that is proved.

\begin{figure}[t]
\begin{lstlisting}
structure Grant where          -- trimmed; the full record has more axes
  spawn    : SpawnTier         -- off < drive < persist
  shell    : Bool
  redshift : Bool
  mailTo   : Cover := []
  git      : GitGrant
  web      : WebTier

structure BudgetFile where
  fuel  : Nat                  -- the limit the parent lent
  spent : Nat := 0             -- how much of it is used

theorem grant_restrict_le (parent role : Grant) :
    (parent.restrict role).le parent = true
\end{lstlisting}
\caption{A grant is a record of axes, a child's grant is a meet with its parent's, and the bound is a theorem.}
\label{lst:grant}
\end{figure}

\subsection{Reversibility and review}
\label{sec:reversibility}

\paragraph{Tiers.} Every effect class declares an undo ceiling (\Cref{tab:caps}). From strongest to weakest, its
effects are:
\begin{itemize}[leftmargin=1.5em,nosep]
  \item read-only: they change nothing;
  \item reversible: the harness can restore what was there, on this host or across hosts;
  \item agent-reversible: the harness can restore the agent's own state, though the world may already have seen
    the effect;
  \item irreversible.
\end{itemize}
A tool's ceiling is the weakest of the ceilings of the classes in its footprint, and its tier follows from that
ceiling: \emph{jail} for read-only, \emph{tame} for reversible, \emph{feral} for agent-reversible, and \emph{wild} for
irreversible. One rule then lowers the tier. When a token bounds every class in the footprint, the tool is tame
whatever its ceiling, because the human already approved its reach when they scoped the token. So
\texttt{File\_write} is tame because the harness can undo it, and \texttt{Git\_push} and \texttt{Web\_fetch} are tame
because their tokens name the remotes and hosts they may reach, though neither can be undone. \texttt{Shell\_run} is
neither undoable nor bounded, so it is wild and, under the default review posture, every call is reviewed
(\Cref{tab:claims}). The rule only ever lowers
a tier (\lean{tier\_le\_default}), so tame means undoable \emph{or} bounded in advance. A tool's tier is computed from its
classes and never declared by the tool. A tool may declare itself less undoable than its classes allow, but not more, and a
build that contains a tool that over-claims refuses to boot. 

\paragraph{Undo.} Before a file tool writes, the harness saves the file's old bytes to Git's object store. At the end
of the turn it seals that turn's saves into one commit on the agent's undo stream, and the undo journal records the
branch head and the stream's tip for that turn. Rewinding to a turn restores every file changed since, from its saved
bytes, and resets the branches (\Cref{sec:lifecycle}). Both round trips are theorems under \emph{named assumptions}: a named assumption is a
Lean proposition about the world, stated with the observations that would falsify it, and a theorem that needs one
takes it as an explicit hypothesis, so the assumption shows in the theorem's type. A saved
file reads back as the bytes it replaced, if Git's object store returns what it hashed (\Cref{tab:claims}). A commit
is undone by resetting to the old head, if \texttt{git reset} restores a ref
(\lean{live\_commit\_reset\_round\_trips}). Saved bytes last only as long as the retention window before Git's garbage
collection. What undo cannot reach is whatever has left the host, such as a push, a request to a web host or a
command that already ran.

\paragraph{Evidence for a ceiling.} A class's ceiling is declared, and this is one place where the formalization is
incremental (\Cref{app:formalization}). Beside each declaration the harness records why it is honest, typed against
the ceiling it supports (\Cref{lst:reversible}). An irreversible ceiling needs no evidence, because it is the weakest
claim and always safe. A read-only class names the check that its operations write nothing and the assumption
that their host implementations change no state; these are references, not proof terms. A reversible or
agent-reversible class carries a round trip: an instance of \lean{Reversible}, which cannot be built without a proof
that undoing an applied operation restores exactly the state before it, and an agent-reversible one also names what
the world may already have observed. The file class cites the round trip of one anchored write, whose undo expects
the new bytes and puts back the old ones; the theorem of the previous paragraph ties that model to the live store.
The evidence decides nothing at run time. The tier reads only the ceiling, so a proof moves no tier; it turns a
claim into a theorem. Proving an abstract inverse and showing that a provider implements it remain two obligations,
and the second rests on the provider, which belongs to the trusted base.

\begin{figure}[ht]
\begin{lstlisting}
class Reversible (Op State Err : Type) where     -- an undo, with its proof
  apply    : Op -> State -> Except Err State      -- refuses if the state has drifted
  produces : Op -> State                          -- the state an applied op leaves
  reverse  : Op -> Op
  restores : forall o s, apply o s = .ok (produces o) ->
               apply (reverse o) (produces o) = .ok s

structure CellOp (a : Type) where                -- one anchored write
  pre  : Option a                                 -- what it expects; none = absent
  post : Option a                                 -- what it leaves; none = deleted
-- reverse swaps pre and post; restores is a two-line proof

inductive CeilingEvidence : ReversalClass -> Type 1 where
  | bottom    (what : String)            : CeilingEvidence (.irreversible what)
  | frame     (spelling world : String)  : CeilingEvidence .readOnly
  | roundTrip (inst : Reversible Op S E) (s : RestoreScope) : CeilingEvidence (.reversible s)
  | hostRoundTrip (inst : Reversible Op S E) (s : RestoreScope) (w : String)
      : CeilingEvidence (.agentReversible s w)

instance : Ceiling "FsWriteEffects" where
  evidence := .roundTrip (instReversibleCell (a := String)) .sameHost
\end{lstlisting}
\caption{How a ceiling is known. An undo is a \lean{Reversible} instance, which carries its proof; each class's
evidence is typed by the ceiling it supports; and the file class cites the round trip of one anchored write. The
law that reversing twice is the identity is omitted.}
\label{lst:reversible}
\end{figure}

\paragraph{Review.} A reviewed call goes through a short, bounded exchange before it runs, in place of a permission
dialog:
\begin{enumerate}[leftmargin=1.5em,nosep]
  \item A \emph{reviewer} model reads the tool, its arguments and the agent's stated reason, and returns
    \texttt{safe} or \texttt{wild}, optionally naming a tame alternative. By default the reviewer is a smaller model,
    or the agent's own model when no smaller one is configured.
  \item Every verdict, \texttt{safe} included, goes back to the proposing agent, which \emph{keeps} the call,
    \emph{redirects} to another way of doing the work, or \emph{withdraws} it. Keeping a \texttt{safe} call runs it; a
    redirect or a withdrawal means it never runs. An agent that asks to run \texttt{grep -rn TODO src/} through
    \texttt{Shell\_run}, for example, may be pointed to \texttt{File\_grep}, which does the same work unreviewed.
  \item Keeping a \texttt{wild} call sends the agent's defence back to the reviewer, by default for one more round.
    A disagreement that survives it is a \emph{stalemate}, and only there is the human asked: the card shows the call,
    the agent's reason and the reviewer's verdict.
\end{enumerate}
A call runs only through a \texttt{safe} verdict the agent kept, or a stalemate the human approved
(\lean{run\_needs\_safe\_or\_yes}), and every exchange ends in a bounded number of steps (\lean{trace\_bounded}). With
no human present, the stalemate's answer is given in advance: \texttt{-{}-no}, the default for unattended sessions,
rejects, and \texttt{-{}-yes} approves. The verdict is read only from the first line that begins \texttt{VERDICT:}; a
reply without one counts as \texttt{wild} (\lean{debate\_no\_verdict\_is\_wild}), and so does a reviewer that cannot be
reached, and tool calls in the reviewer's reply are ignored. Review is defence in depth, not a boundary. The tokens
are the boundary, and a human can still approve a bad call. A reviewer is also fallible in ways the exchange cannot
see: a model-written reason can omit where a value came from, or what private data an otherwise permitted request
carries, and a bounded destination does not make every disclosure to it acceptable.

\paragraph{Vigils.} An agent that must wait, for a build, a remote branch or a letter, arms a \emph{vigil}: a
condition the harness checks on the agent's behalf. When the condition becomes true, the agent wakes with a turn. Each
vigil has its own budget of wakes, and a model may arm at most sixteen. A parent can \emph{pin} a vigil on a child by
writing it beside the child's grant. The child's model cannot cancel, replace or retune a pinned vigil
(\lean{locked\_invariant\_under\_model\_verbs}), and reloading the parent's pins never touches the vigils the child
armed itself. For example, a planner can pin ``wake when \texttt{mainline} moves'' on each worker, so that every
worker hears of every landing. The guarantee is that the child's model cannot remove the pin; delivery still
depends on the daemon and on the vigil's budget of wakes, and nothing says the child acts on it.

\subsection{Self-extension: sealed programs}
\label{sec:sealed}

An agent extends its own tool surface by writing a Lean program. A file \texttt{agent/programs/todos.lean} on its
persona branch, say one that counts the \texttt{TODO} comments under \texttt{src/}, becomes the tool
\texttt{Program\_todos} at the agent's next boot. Writing the file is an ordinary, tame edit. What makes the program
safe to run is what the harness checks when it loads it (\Cref{lst:program}).

\begin{figure}[t]
\begin{lstlisting}
structure ToolProgram where    -- trimmed
  description : String
  footprint   : List String    -- the classes the program declares
  run : {m : Type -> Type} -> [Monad m]
      -> [FsReadEffects m] -> [FsWriteEffects m] -> [GitReadEffects m]
      -> [GitLocalEffects m] -> [GitFetchEffects m] -> [NetEffects m]
      -> [AwsAuthEffects m] -> [AgentEffects m] -> [ClockEffects m]
      -> [SpawnEffects m] -> [KubeReadEffects m]
      -> ProgramCaps -> String -> Confined footprint m String
\end{lstlisting}
\caption{A sealed program's type is its permission request. The body is polymorphic in the host monad, so it can
reach the world only through the eleven classes it is given, and \texttt{Confined} admits only the classes the
program lists in \texttt{footprint}. \texttt{ProgramCaps} holds the tokens the agent's grant minted, including the
child-lifecycle door when the grant may spawn.}
\label{lst:program}
\end{figure}

The harness makes three checks:
\begin{itemize}[leftmargin=1.5em,nosep]
  \item \textbf{The seal.} The file must be a single term, with no tactics, options or local recursion. This shape
    is checked on the macro-expanded syntax \emph{before} elaboration, because elaborating source can itself run code,
    and rejecting the result afterwards could not undo that. Every
    constant in its closure must be safe: nothing \texttt{unsafe}, \texttt{partial} or implemented natively, no
    token constructed outside its module, and no axiom beyond Lean's standard three. The last condition also rules
    out \texttt{sorry} (\Cref{tab:claims}). The program is elaborated in a restricted vocabulary that contains no
    token constructor and no bridge to raw IO.
  \item \textbf{The type.} The term must be a \lean{ToolProgram}. Its body is written over an unknown host monad, so
    it can do only what the eleven classes it is given allow. It can use only the ones it declares, because
    \lean{Confined} provides a class's methods only when the footprint lists the class. So a program that fetches a
    URL without declaring \lean{NetEffects} does not type-check.
  \item \textbf{The surface.} The footprint must lie within the classes admitted for programs. The program is
    offered to the model only if the agent's tokens cover it, by the same check as a built-in tool
    (\Cref{sec:tokens}).
\end{itemize}
At run time a program receives the tokens the grant minted, never the grant itself. It has no interface for calling
the model, building, running a shell or waiting on a vigil. It can create, drive and retire children, but through
the same spawn gate as the tool, so each child is bounded by the agent's grant. It can therefore reach only what its
author could already reach. Its tier follows from its footprint,
like any tool's. The \texttt{TODO} counter names only \lean{FsReadEffects}, so it runs unreviewed. A program
that signs requests to AWS is reviewed, because that class has no undo ceiling and an unknown class counts as wild.
What is proved is local: an admitted term's syntax contains no refused node kind
(\lean{shape\_admitted\_has\_no\_refused\_kind}), and every constant it names directly is admitted
(\lean{sealUsed\_admits\_only\_safe}). That these checks, with the trusted elaborator and the providers, add up to
full confinement of a malicious source is argued by construction and pinned by tests, not proved
(\Cref{app:formalization}).

Hooks (\Cref{sec:personas}) are sealed the same way, but a hook only returns a disposition, which the harness carries
out. A plugin is a bundle of programs. If any one of its programs fails a check, the whole plugin is refused, and a
loaded plugin's name and declared footprint are within what its admission allows
(\lean{loaded\_plugin\_authority\_le\_cap}).

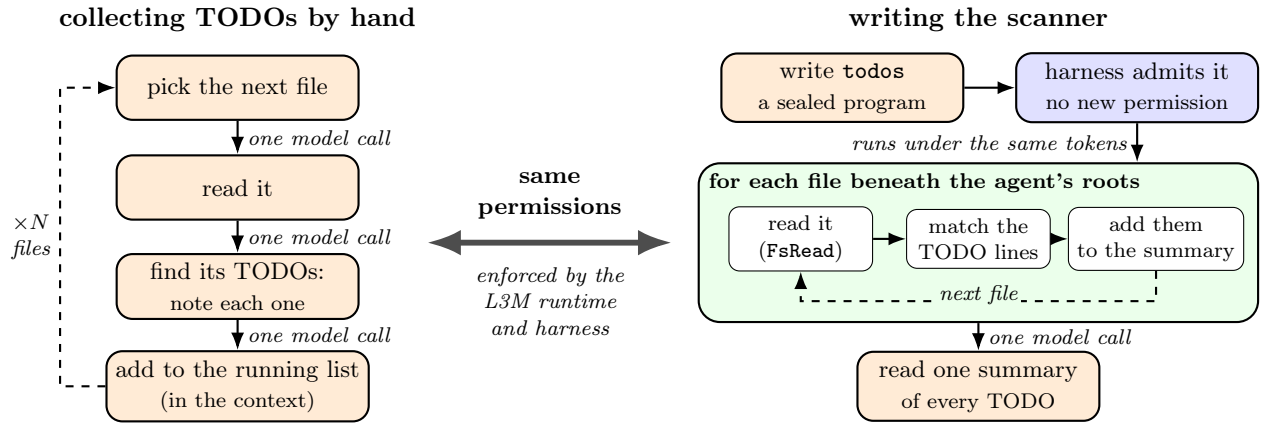
\begin{figure}[ht]
\centering
\begin{tikzpicture}[
  font=\small, >=Latex,
  mb/.style={draw, thick, rounded corners=6pt, fill=orange!16, minimum width=3.2cm, minimum height=0.85cm, align=center},
  cb/.style={draw, thick, rounded corners=6pt, fill=green!14, minimum width=3.2cm, minimum height=0.85cm, align=center},
  hb/.style={draw, thick, rounded corners=6pt, fill=blue!12, minimum width=3.2cm, minimum height=0.85cm, align=center},
  sb/.style={draw, rounded corners=4pt, fill=white, minimum width=2.0cm, minimum height=0.7cm, align=center, font=\footnotesize},
  arc/.style={->, thick},
  alab/.style={font=\footnotesize\itshape, align=center},
  title/.style={font=\normalsize\bfseries}
]
  \node[title] at (-4.6,3.4) {collecting TODOs by hand};
  \node[mb] (l1) at (-4.6,2.5) {pick the next file};
  \node[mb] (l2) at (-4.6,1.18) {read it};
  \node[mb] (l3) at (-4.6,-0.13) {find its TODOs:\\ \footnotesize note each one};
  \node[mb] (l4) at (-4.6,-1.45) {add to the running list\\ \footnotesize (in the context)};
  \draw[arc] (l1) -- node[alab, right] {one model call} (l2);
  \draw[arc] (l2) -- node[alab, right] {one model call} (l3);
  \draw[arc] (l3) -- node[alab, right] {one model call} (l4);
  \draw[arc, dashed] (l4.west) -- ++(-0.6,0) |- node[alab, left, pos=0.25] {$\times N$\\ \footnotesize files} (l1.west);

  \draw[<->, line width=2.2pt, gray!60!black] (-2.1,0.45) -- (1.15,0.45);
  \node[align=center, font=\small\bfseries, anchor=south] at (-0.47,0.65) {same\\ permissions};
  \node[align=center, font=\footnotesize\itshape, anchor=north] at (-0.47,0.25) {enforced by the\\ L3M runtime\\ and harness};

  \node[title] at (5.2,3.4) {writing the scanner};
  \node[mb] (r1) at (3.4,2.5) {write \texttt{todos}\\ \footnotesize a sealed program};
  \node[hb] (r2) at (7.3,2.5) {harness admits it\\ \footnotesize no new permission};
  \node[draw, thick, rounded corners=8pt, fill=green!8, minimum width=7.4cm, minimum height=2.1cm] (loop) at (5.2,0.45) {};
  \node[anchor=north west, font=\footnotesize\bfseries, inner sep=4pt] at (loop.north west) {for each file beneath the agent's roots};
  \node[sb, minimum width=1.9cm] (b1) at (2.85,0.5) {read it\\ \footnotesize (\texttt{FsRead})};
  \node[sb, minimum width=1.9cm] (b2) at (5.2,0.5) {match the\\ TODO lines};
  \node[sb, minimum width=1.9cm] (b3) at (7.55,0.5) {add them\\ to the summary};
  \draw[arc] (b1) -- (b2); \draw[arc] (b2) -- (b3);
  \draw[arc, dashed] (b3.south) -- ++(0,-0.4) -| (b1.south);
  \node[alab, fill=green!8, inner sep=1pt] at (5.2,-0.24) {next file};
  \node[mb] (r4) at (5.2,-1.45) {read one summary\\ \footnotesize of every TODO};
  \draw[arc] (r1) -- (r2);
  \draw[arc] (r2.south) -- node[alab, left] {runs under the same tokens} (r2.south |- loop.north);
  \draw[arc] (loop.south) -- node[alab, right] {one model call} (r4.north);
\end{tikzpicture}
\caption{Two ways to collect the \texttt{TODO}s in a tree. Left: the agent is the loop, and every file costs model
calls and context. Right: the agent writes the scan as a sealed program that runs under the tokens it already holds;
each file is read and matched with no model call, and the model reasons once, over the summary.}
\label{fig:workflow}
\end{figure}

The result is a programming model (\Cref{fig:workflow}). Because the bound is carried into every call, an
agent may write a new tool as a sealed program and have it join its own tool surface with no new permission:
the tool reaches exactly the capabilities the agent already held, and every invariant that held before still
holds. Capability grows; authority does not. The model does what it is good at --- deciding what to do and
composing the pieces --- and verified code does what it is good at --- computing and enforcing. Because tame
actions are reversible the agent can act without hedging, and because review is a deliberation between the
proposing agent and a reviewer the human is consulted only at a stalemate rather than at every step. We argue that
this is a building block for long-running, many-agent systems: the alternatives are an agent that asks
forever or one that is trusted blindly.

\section{Multi-Agent Architecture: Putting It Together}
\label{sec:arch}

\Cref{sec:agents} described one agent. This section describes many: where they run (\Cref{sec:hubs}), how authority
narrows as their tree grows (\Cref{sec:tree}), and how they talk (\Cref{sec:mail}).

\subsection{Hubs and the daemon}
\label{sec:hubs}

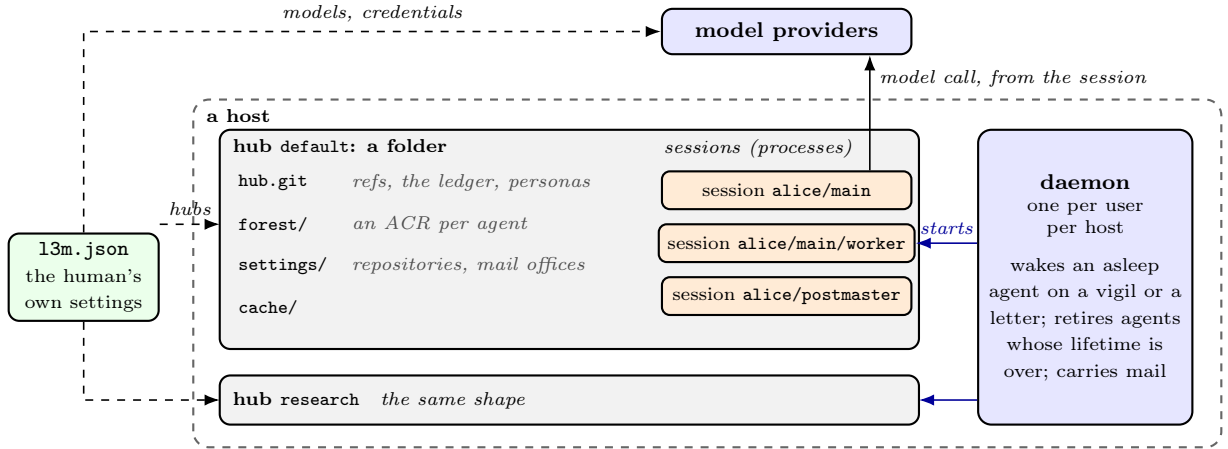
\begin{figure}[t]
\centering
\begin{tikzpicture}[
  font=\footnotesize, >=Latex,
  every node/.append style={execute at begin node={\hyphenpenalty=10000\relax}},
  f/.style={font=\scriptsize\ttfamily, anchor=west, inner sep=1.2pt},
  fd/.style={font=\scriptsize\itshape, anchor=west, inner sep=1.2pt, text=black!70},
  sess/.style={draw, thick, rounded corners=3pt, fill=orange!16, minimum width=3.3cm, minimum height=0.5cm,
               font=\scriptsize},
  lab/.style={font=\scriptsize\itshape, align=center},
]
  \draw[thick, dashed, rounded corners=6pt, black!60] (2.4,-3.3) rectangle (16.0,1.3);
  \node[anchor=north west, font=\scriptsize\bfseries] at (2.45,1.3) {a host};

  \draw[thick, rounded corners=4pt, fill=gray!10] (2.75,-2.0) rectangle (12.0,0.9);
  \node[anchor=north west, font=\scriptsize\bfseries] at (2.8,0.9) {hub \texttt{default}: a folder};
  \node[f] (r1) at (2.95,0.2)   {hub.git};   \node[fd] at (4.45,0.2)   {refs, the ledger, personas};
  \node[f] (r2) at (2.95,-0.35) {forest/};   \node[fd] at (4.45,-0.35) {an ACR per agent};
  \node[f] (r3) at (2.95,-0.9)  {settings/}; \node[fd] at (4.45,-0.9)  {repositories, mail offices};
  \node[f] (r4) at (2.95,-1.45) {cache/};

  \node[anchor=north west, font=\scriptsize\itshape] at (8.5,0.9) {sessions (processes)};
  \node[sess] (s1) at (10.25,0.1)   {session \texttt{alice/main}};
  \node[sess] (s2) at (10.25,-0.6)  {session \texttt{alice/main/worker}};
  \node[sess] (s3) at (10.25,-1.3)  {session \texttt{alice/postmaster}};

  \draw[thick, rounded corners=4pt, fill=gray!10] (2.75,-3.0) rectangle (12.0,-2.35);
  \node[anchor=west, font=\scriptsize] (h2) at (2.8,-2.675) {\textbf{hub \texttt{research}}\quad\textit{the same shape}};

  \node[draw, thick, rounded corners=5pt, fill=blue!10, align=center, text width=2.6cm, minimum height=3.9cm]
    (d) at (14.2,-1.05) {\textbf{daemon}\\ \scriptsize one per user per host\\[4pt]
    \scriptsize wakes an asleep agent on a vigil or a letter; retires agents whose lifetime is over; carries mail};
  \draw[->, semithick, blue!60!black] (d.west |- s2) -- node[lab, above] {starts} (s2.east);
  \draw[->, semithick, blue!60!black] (d.west |- 0,-2.675) -- (12.0,-2.675);

  \node[draw, thick, rounded corners=4pt, fill=blue!10, minimum width=3.3cm, minimum height=0.6cm] (prov)
    at (10.25,2.2) {\textbf{model providers}};
  \draw[->, semithick] ([xshift=1.1cm]s1.north) -- node[lab, right, pos=0.8] {model call, from the session}
    ([xshift=1.1cm]prov.south);

  \node[draw, thick, rounded corners=4pt, fill=green!8, align=center, text width=1.75cm] (u) at (0.95,-1.05)
    {\texttt{l3m.json}\\ \scriptsize the human's own settings};
  \draw[->, semithick, dashed] (u.east |- 0,-0.35) -- node[lab, above] {hubs} (2.75,-0.35);
  \draw[->, semithick, dashed] (u.south) |- (2.75,-2.675);
  \draw[->, semithick, dashed] (u.north) |- node[lab, above, pos=0.75] {models, credentials} (prov.west);
\end{tikzpicture}
\caption{L3M on one host. One daemon serves every hub on the host, and a hub is a folder: a bare repository, a forest
of ACRs, its settings and a cache. Sessions are processes that the daemon starts and that call the model themselves.
The human's own settings name the hubs, the models and the credentials.}
\label{fig:arch}
\end{figure}

A hub is a folder with no process of its own (\Cref{fig:arch}). It holds \texttt{hub.git}; a \texttt{forest/} of ACRs
nested by name, so the body of \texttt{alice/main/worker} is under \texttt{forest/alice/main/worker/self/}; a
\texttt{settings/} checkout of the hub's settings branch, which lists the repositories the hub knows, its mail offices
and its terminal layout; and a cache. The first session creates a default hub under
\texttt{\textasciitilde/.l3m/hub/}. A human may keep more, and a session's hub is the one its ACR is in. The human's
own document, \texttt{\textasciitilde/.l3m/l3m.json}, names the models, the credentials, the hubs and the hosts. It is
the only source of credentials, and like the hub's settings it is written only by the human (\Cref{sec:threat}).

One daemon runs per user per host and serves every hub on it. It restarts an agent that is asleep when one of its
vigils fires or a letter arrives: an agent with no human attached leaves when it goes idle, and the daemon wakes only
agents that are asleep (\lean{wakes\_iff\_asleep}). It also retires agents whose lifetime is over
(\lean{lifetimeRetires\_iff\_done}) and carries mail between hubs (\Cref{sec:mail}). No decision of a session depends
on it: the daemon starts sessions and moves letters, and every check runs inside the session. The model call is the
session's own as well (\Cref{sec:log}), made with credentials from the human's document alone. The human can also list
other hosts, whose hubs the daemon reports beside the local ones; what travels between hosts is
status and views, never authority.

\subsection{Delegation and the agent tree}
\label{sec:tree}

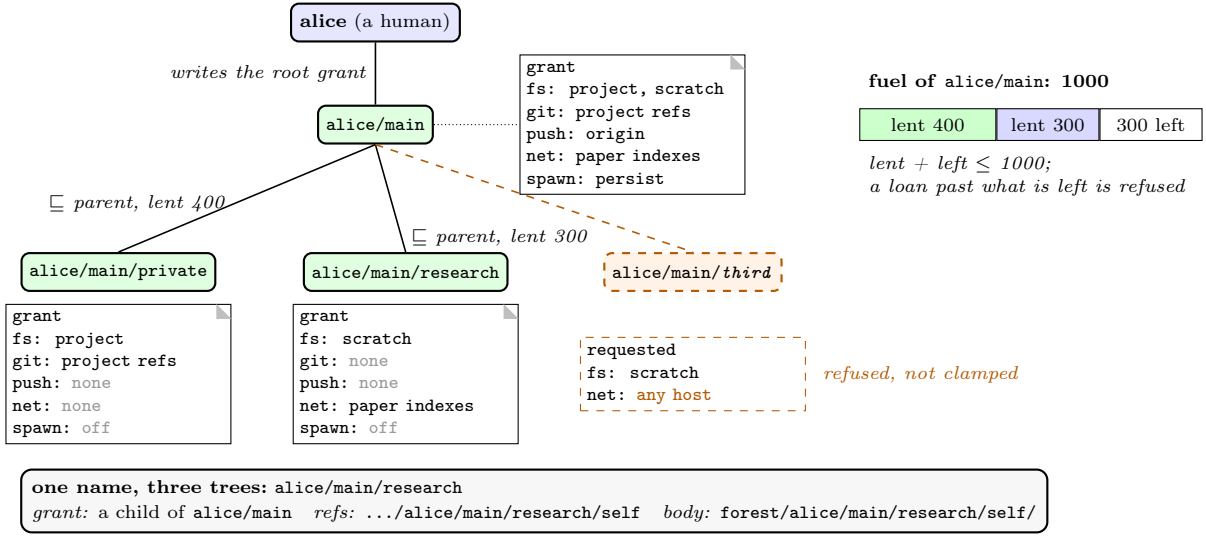
\begin{figure}[t]
\centering
\begin{tikzpicture}[
  font=\scriptsize, >=Latex,
  every node/.append style={execute at begin node={\hyphenpenalty=10000\relax}},
  ag/.style={draw, thick, rounded corners=3pt, fill=green!12, font=\scriptsize\ttfamily, minimum height=0.5cm,
             inner sep=3pt},
  grant/.style={draw, fill=white, font=\scriptsize\ttfamily, align=left, inner sep=2.5pt, text width=2.8cm},
  lab/.style={font=\scriptsize\itshape, align=center},
  fuel/.style={draw, minimum height=0.42cm, inner sep=0pt, font=\scriptsize},
]
  \newcommand{\none}[1]{\textcolor{black!40}{#1}}
  \newcommand{\gcard}[3]{
    \node[grant] (#1) at #2 {\textbf{grant}\\ #3};
    \fill[black!25] ($(#1.north east)+(-0.2,0)$) -- ++(0.2,-0.2) -- ++(-0.2,0) -- cycle;}
  \node[draw, thick, rounded corners=3pt, fill=blue!10, minimum height=0.5cm] (h) at (4.9,3.1)
    {\textbf{alice} (a human)};
  \node[ag] (P) at (4.9,1.75) {alice/main};
  \draw[semithick] (h) -- node[lab, left] {writes the root grant} (P);
  \gcard{gP}{(8.3,1.75)}{fs:\ \ project, scratch\\ git: project refs\\ push: origin\\ net: paper indexes\\ spawn: persist}
  \draw[densely dotted] (P.east) -- (gP.west |- P.east);

  \node[anchor=south west, font=\scriptsize\bfseries] at (11.3,2.1) {fuel of \texttt{alice/main}: 1000};
  \node[fuel, fill=green!20, minimum width=1.8cm, anchor=west] (f1) at (11.3,1.75) {lent 400};
  \node[fuel, fill=blue!15,  minimum width=1.35cm, anchor=west] (f2) at (f1.east) {lent 300};
  \node[fuel, fill=white,    minimum width=1.35cm, anchor=west] (f3) at (f2.east) {300 left};
  \node[lab, anchor=north west, align=left] at (11.3,1.45) {lent + left $\le$ 1000;\\ a loan past what is left is refused};

  \node[ag] (C1) at (1.5,-0.2) {alice/main/private};
  \node[ag] (C2) at (5.3,-0.2) {alice/main/research};
  \node[ag, dashed, draw=orange!70!black, fill=orange!10] (C3) at (9.1,-0.2) {alice/main/\textit{third}};
  \draw[semithick] (P.south) -- node[lab, left, pos=0.55] {$\sqsubseteq$ parent, lent 400} (C1.north);
  \draw[semithick] (P.south) -- node[lab, right, pos=0.86] {$\sqsubseteq$ parent, lent 300} (C2.north);
  \draw[semithick, dashed, orange!70!black] (P.south) -- (C3.north);
  \gcard{g1}{(1.5,-1.55)}{fs:\ \ project\\ git: project refs\\ push: \none{none}\\ net: \none{none}\\ spawn: \none{off}}
  \gcard{g2}{(5.3,-1.55)}{fs:\ \ scratch\\ git: \none{none}\\ push: \none{none}\\ net: paper indexes\\ spawn: \none{off}}
  \node[grant, draw=orange!70!black, dashed] (g3) at (9.1,-1.55)
    {\textbf{requested}\\ fs:\ \ scratch\\ net: \textcolor{orange!70!black}{any host}};
  \node[lab, text=orange!70!black, anchor=west] at ($(g3.east)+(0.1,0)$) {refused, not clamped};

  \node[draw, thick, rounded corners=4pt, fill=gray!6, inner sep=4pt, anchor=north west, align=left] at (0.2,-2.8)
    {\textbf{one name, three trees:}\ \texttt{alice/main/research}\\[1pt]
     \textit{grant:} a child of \texttt{alice/main}\quad
     \textit{refs:} \texttt{\dots/alice/main/research/self}\quad
     \textit{body:} \texttt{forest/alice/main/research/self/}};
\end{tikzpicture}
\caption{A tree of agents at spawn time. Each child's grant lies within its parent's, and its fuel is lent from the parent's. A
request that exceeds either is refused. The private child keeps the project's refs but not the right to push
them. The two children are the peers of \Cref{sec:ex-split}.}
\label{fig:grant-tree}
\end{figure}

A child exists because its parent asked the harness to spawn it under a requested grant (\Cref{fig:grant-tree}). The
harness re-reads the parent's grant from the ledger and admits the request only if it lies within that grant; a
widening is refused, never clamped (\lean{spawn\_persist\_within\_parent}, \lean{issueGrant\_refuses\_axis\_gain}). A
human's grant is a root that no agent can write (\lean{human\_grant\_is\_a\_root}). The right to spawn is itself an
axis with three levels, \texttt{off}, \texttt{drive} and \texttt{persist}: a \texttt{drive} parent may start only
children that are retired when the call that started them returns (\lean{drive\_plans\_are\_driven\_retireOnExit}).
Fuel is lent the same way. A loan larger than what the parent has left is refused (\lean{spawn\_refuses\_overdraft}),
and a retired child's loan returns to its parent, whose account is rebuilt from its live children. A child's name is
under its parent's (\lean{descendant\_uri\_under\_parent}), so its refs and its ACR are under the parent's too, and a
parent reads its descendants through one prefix (\Cref{sec:improve}).

Each of these checks is proved at the hop where it is made, and the hops compose. Grants change after spawn: an
agent's grant to another entity is checked against the \emph{issuer}, which may be a grandparent rather than the
parent. So the bound over the whole tree is stated over every ledger the agents themselves can write, and it says
that a descendant's grant lies within its root's grant, extended at each ancestor by that ancestor's own refs and
mail (\lean{descendant\_bounded\_by\_root\_grant}). The extension is necessary: the plain bound by the root's stored
grant is false one level down, because a child is granted its parent's own refs, which the parent's stored grant
does not list. The theorem does not bound a descendant by every ancestor, and it stops where a human's regrant
enters the ledger. The budget bound, by contrast, covers one loan batch at a time (\Cref{app:formalization}).
Retiring an agent retires its subtree (\Cref{sec:lifecycle}). Because each child also gets its own branches and
worktrees, workers develop changes without sharing a branch head, and a parent can compare what they produced.

\subsection{Coordination and mail}
\label{sec:mail}

A letter's address names up to five parties: a worktree, a persona (the agent's name), a branch, a repository and
a human (\Cref{lst:envelope}). Exactly one is starred, the party that must handle the letter; the others are hints,
and any may be blank. The star may land only where the machine can vouch for a reader, on an agent, which the ledger
names, or on a registered human (\lean{star\_admitted\_is\_vouchable}). So \texttt{alice/main/worker} is short for
\texttt{agent:alice/main/worker*}, and \texttt{human:bob* branch:fix} reaches bob's box, which bob's postmaster serves
and routes onward, with the branch as a hint to where it belongs. The sender's own address travels with the letter
under the rule \emph{the sender asserts, the machine refutes}: the harness overwrites the fields it owns, the worktree
and the agent (\lean{stampFrom\_machine\_owned}); it refuses a human it knows to be wrong; and it passes a branch or
repository as the hint it is (\lean{refute\_blind\_on\_hints}). A letter to another hub is prefixed with an office, as
in \texttt{office:bob}. Whom an agent may write to is an axis of its grant. By default it covers the agent's own
subtree and each ancestor's inbox, so siblings cannot mail each other unless a grant says so, and delivery goes
through an inbox token that appends only to the inboxes that axis covers (\lean{inboxCap\_write\_covered}).

\begin{figure}[ht]
\begin{minipage}[t]{0.5\linewidth}
\begin{lstlisting}[xleftmargin=0.5em]
structure Envelope where    -- one line of an inbox
  id      : MessageId       -- host
  ts      : Int             -- host
  sender  : Sender          -- host; written "from"
  orig    : Option String   -- host: sender's address
  re      : Option MessageId -- reply tool
  addr    : Option String   -- the address, as written
  body    : String          -- model
  title   : Option String   -- model: a subject
  display : Option String   -- model: for the human
\end{lstlisting}
\end{minipage}%
\begin{minipage}[t]{0.5\linewidth}
\begin{lstlisting}[xleftmargin=0.5em]
structure Tuple where       -- each a hint; may be blank
  worktree persona branch : Option String
  repo human : Option String

structure Address where     -- "human:bob* branch:fix"
  tuple : Tuple
  star  : Field             -- must handle it

structure Stamped where     -- a letter for another hub
  office : String
  to     : String
  env    : Envelope
\end{lstlisting}
\end{minipage}
\caption{A letter, and its address. The host writes every field that says who sent a letter and when; the model
writes the rest. An address names five parties, the persona being the agent's name, and stars one. A letter for
another hub adds the office and the recipient.}
\label{lst:envelope}
\end{figure}

A letter is an envelope around the model's body (\Cref{lst:envelope}). The host, not the sender, stamps its id, time
and sender; the send tool has no \texttt{from} argument. The model may add a subject and a one-line summary for the
human, and a reply carries the id of the letter it answers, which the reply tool stamps. A letter
is one line that cannot contain a newline (\lean{render\_no\_newline}), so a body cannot pose as a second letter. An
inbox is an append-only file on the recipient's inbox ref, and concurrent appends merge by set union, which loses no
letter and invents none (\lean{merge\_no\_fabrication}). Reading a letter appends
a read mark, which is the letter's receipt and on which its sender may arm a vigil. An arriving letter wakes its
recipient, and the recipient's hook decides what the model sees (\Cref{sec:personas}). L3M defines no protocol above
this: a request and its reply are a letter and its answer, completion is a receipt or a letter to the parent, and a
box that stops draining is reported to the postmaster.

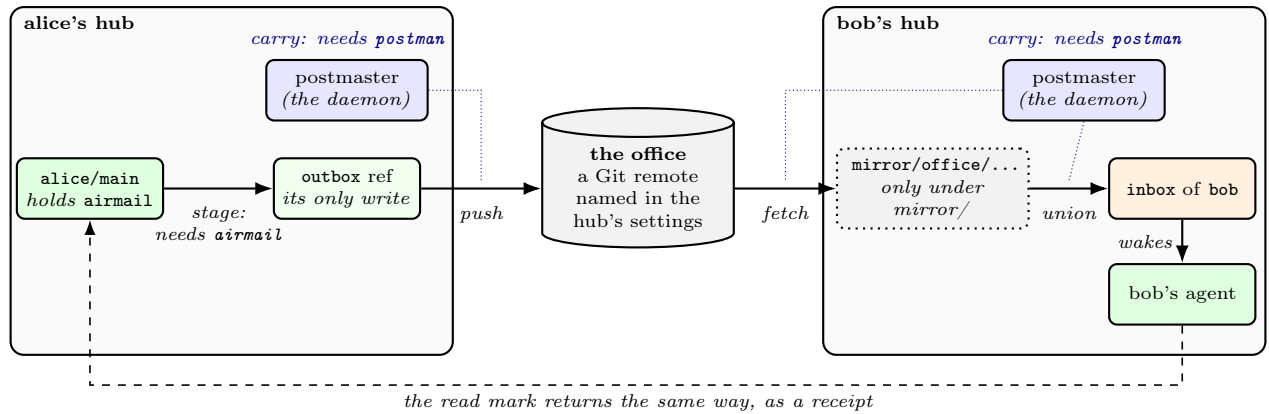
\begin{figure}[t]
\centering
\begin{tikzpicture}[
  font=\scriptsize, >=Latex,
  every node/.append style={execute at begin node={\hyphenpenalty=10000\relax}},
  box/.style={draw, thick, rounded corners=3pt, align=center, minimum height=0.8cm, text width=1.9cm},
  pm/.style={box, fill=blue!10},
  lab/.style={font=\scriptsize\itshape, align=center},
  flow/.style={->, thick},
]
  \draw[thick, rounded corners=5pt, fill=gray!4] (-0.3,-2.2) rectangle (5.55,2.4);
  \node[anchor=north west, font=\scriptsize\bfseries] at (-0.25,2.4) {alice's hub};
  \draw[thick, rounded corners=5pt, fill=gray!4] (10.45,-2.2) rectangle (16.3,2.4);
  \node[anchor=north west, font=\scriptsize\bfseries] at (10.5,2.4) {bob's hub};

  \node[box, fill=green!12, text width=1.7cm] (a) at (0.75,0) {\texttt{alice/main}\\ \textit{holds} \texttt{airmail}};
  \node[box, fill=green!6, text width=1.7cm] (o) at (4.15,0) {\texttt{outbox} ref\\ \textit{its only write}};
  \node[draw, thick, cylinder, shape border rotate=90, aspect=0.15, fill=gray!10, align=center, text width=2.3cm,
        minimum height=1.3cm] (off) at (8.0,0) {\textbf{the office}\\ a Git remote named in the hub's settings};
  \node[box, fill=gray!10, dotted, text width=2.3cm] (m) at (11.9,0) {\texttt{mirror/office/\dots}\\ \textit{only under mirror/}};
  \node[box, fill=orange!12, text width=1.7cm] (i) at (15.2,0) {\texttt{inbox} of \texttt{bob}};
  \node[box, fill=green!12, text width=1.7cm] (b) at (15.2,-1.4) {bob's agent};

  \draw[flow] (a) -- node[lab, below=3pt] {stage:\\ needs \texttt{airmail}} (o);
  \draw[flow] (o) -- node[lab, below=3pt] {push} (off);
  \draw[flow] (off) -- node[lab, below=3pt] {fetch} (m);
  \draw[flow] (m) -- node[lab, below=3pt] {union} (i);
  \draw[flow] (i) -- node[lab, left] {wakes} (b);

  \node[pm] (pa) at (4.15,1.3) {postmaster\\ \textit{(the daemon)}};
  \node[pm] (pb) at (13.9,1.3) {postmaster\\ \textit{(the daemon)}};
  \draw[densely dotted, blue!60!black] (pa.east) -| ($(o.east)!0.5!(off.west)+(0,0.1)$);
  \draw[densely dotted, blue!60!black] (pb.west) -| ($(off.east)!0.5!(m.west)+(0,0.1)$);
  \draw[densely dotted, blue!60!black] (pb.south) -- ($(m.east)!0.5!(i.west)+(0,0.1)$);
  \node[lab, anchor=south, text=blue!50!black] at ($(pa.north)+(0,0.02)$) {carry: needs \texttt{postman}};
  \node[lab, anchor=south, text=blue!50!black] at ($(pb.north)+(0,0.02)$) {carry: needs \texttt{postman}};

  \draw[->, semithick, dashed] (b.south) -- (b.south |- 0,-2.6) -| (a.south);
  \node[lab, below] at (8.0,-2.6) {the read mark returns the same way, as a receipt};
\end{tikzpicture}
\caption{Mail between hubs. The sender's authority ends at its own outbox; carrying mail across machines is the
postmaster's duty, and only the human's settings name the offices. Read marks travel back the same way, as receipts.}
\label{fig:airmail}
\end{figure}

Mail to another hub crosses machines in two steps held by different parties (\Cref{fig:airmail}). Only an agent
holding the \texttt{airmail} axis may stage a letter for another hub, and it stages it on its own outbox ref
(\lean{no\_airmail\_no\_outbox\_write}). Carrying needs the \texttt{postman} axis, which the hub's postmaster holds
(\lean{no\_postman\_no\_office\_authority}). The daemon, acting for the postmaster, pushes outboxes to each
\emph{office}, a Git remote named in the hub's settings; fetches offices only into mirror refs
(\lean{officeFetchDest\_under\_mirror}); relays their letters into inboxes by set union
(\lean{relay\_union\_no\_fabrication}); and carries read marks back as receipts. The sender's authority ends at its own
outbox, and which machines mail can reach is the human's choice.

\section{Separation of Duties}
\label{sec:duties}

Each example in this section splits one piece of work between two holders, so that neither can complete a harmful
action alone. Some of these splits are built into L3M and hold for every agent in every hub: an agent can issue grants
to its children but never to itself (\Cref{tab:claims}), only a human can bless a recipe (\Cref{sec:ex-build}) or
publish a persona (\Cref{sec:personas}), and an agent that stages a letter cannot carry it (\Cref{fig:airmail}).
Others are patterns. Nothing in L3M makes anyone use them, but they are built from the primitives it provides, namely
grants, mail and sealed programs, and each inherits those primitives' guarantees. The builder and runner of
\Cref{sec:ex-build}, the private agent and researcher of \Cref{sec:ex-split}, and the quarantined reader of
\Cref{sec:ex-quarantine} are of this kind.

\subsection{Staged builds: write, gate, execute}
\label{sec:ex-build}

\begin{figure}[t]
\centering
\begin{tikzpicture}[
  font=\scriptsize, >=Latex,
  role/.style={draw, thick, rounded corners=2pt, fill=white, align=center, text width=2.6cm, minimum height=0.8cm,
               inner sep=2pt},
  arr/.style={->, thick},
  lab/.style={font=\scriptsize\itshape},
  inst/.style={font=\scriptsize, text=black!65, align=center, text width=3.4cm},
]
  \node[role] (w) at (0,0) {\textbf{writer}\\ \textit{edits; cannot run}};
  \node[role] (g) at (4.6,0) {\textbf{gate}\\ \textit{pins the bytes by hash}};
  \node[role, text width=3.2cm] (x) at (9.5,0) {\textbf{executor}\\ \textit{runs them read-only;}\\
    \textit{writes only a fresh \texttt{out/}}};
  \draw[arr] (w) -- node[lab, above] {bytes} (g);
  \draw[arr] (g) -- node[lab, above] {pinned bytes} (x);
  \draw[->, thick, dashed, black!55] (w.south) to[bend right=18] node[lab, below, text=black!65] {an edit withdraws the pin} (g.south);
  \node[anchor=east, font=\scriptsize\bfseries, text=black!65] at (-1.8,-1.55) {Make:};
  \node[inst] at (0,-1.55) {an agent edits the Makefile};
  \node[inst] at (4.6,-1.55) {a human blesses it};
  \node[inst] at (9.5,-1.55) {the harness runs them};
\end{tikzpicture}
\caption{Write, gate, execute: W$\oplus$X for authority. No party both writes bytes and runs them, and the
gate's approval names the exact bytes. The staged build of \Cref{fig:staged} is the same shape with two agents.}
\label{fig:wx}
\end{figure}
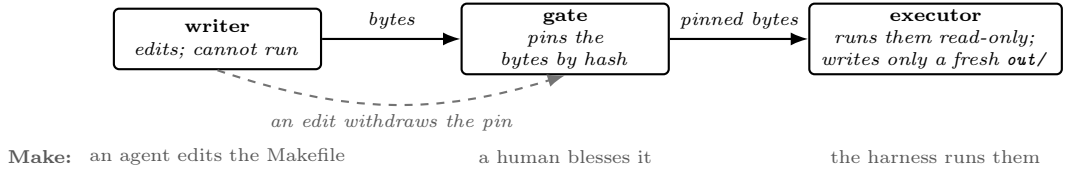

A build is where an agent's writes become code that something executes, so it is where writing and executing meet.
L3M keeps the two apart the way W$\oplus$X keeps a page from being both writable and executable
(\Cref{fig:wx}). A \emph{writer} edits bytes and cannot run them. A \emph{gate} pins the exact bytes it approves by
content hash, so any later edit withdraws the approval. An \emph{executor} runs only the pinned bytes, read-only, and
writes only to a fresh, disposable output directory. A pipeline adds a rule for a second pair, private bytes and the
network: authority only drops along the stages, so once a stage has seen private source no later stage regains the
network, and a dependency found missing is an error in the plan rather than a reason to reconnect.

\textbf{Make.} In the simplest instance the writer is an agent editing a Makefile and the gate is a human.
\texttt{l3m agent bless} is a verb only a human can run: it hashes the Makefile, the files it literally includes and
the scripts its recipes call, and stores the pin on a ref no session can write. A run executes the blessed file only
while those bytes are unchanged and takes no arguments from the model (\lean{blessed\_no\_model\_args}); after an
edit, runs refuse until a human blesses again. Because the blessing is the review, the run shows no review card.
Tame make is the limit case, with no writer left to trust: the harness lowers a restricted Makefile and
writes every command itself, and every output lands under \texttt{build/} (\lean{lower\_recipes\_closed}), so the run
is reversible and needs no human.

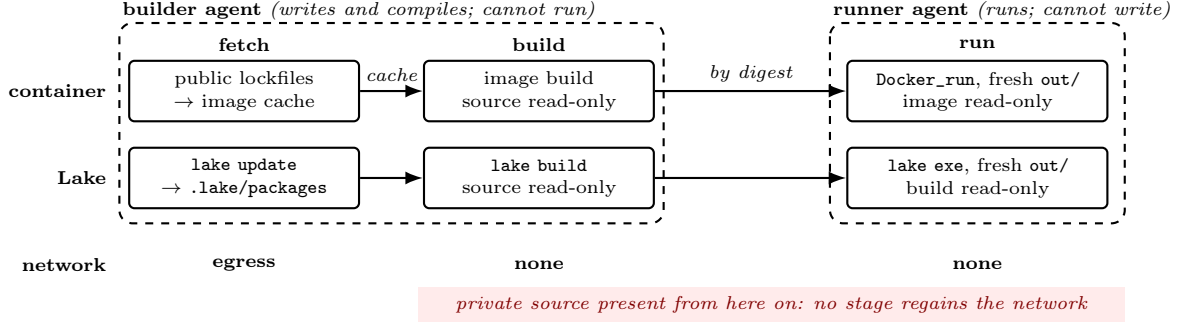
\begin{figure}[t]
\centering
\begin{tikzpicture}[
  font=\scriptsize, >=Latex,
  cell/.style={draw, thick, rounded corners=2pt, fill=white, align=center, text width=2.9cm, minimum height=0.8cm,
               inner sep=2pt},
  arr/.style={->, thick},
  lab/.style={font=\scriptsize\itshape},
  lane/.style={anchor=east, font=\scriptsize\bfseries},
  grp/.style={thick, dashed, rounded corners=5pt},
]
  \fill[red!8] (4.9,-3.05) rectangle (14.25,-2.6);
  \node[font=\scriptsize\itshape, text=red!50!black] at (9.575,-2.825)
    {private source present from here on: no stage regains the network};
  \node[lane] at (0.9,0) {container};
  \node[lane] at (0.9,-1.15) {Lake};
  \node[cell] (cf) at (2.6,0) {public lockfiles\\ $\to$ image cache};
  \node[cell] (cb) at (6.5,0) {image build\\ source read-only};
  \node[cell, text width=3.3cm] (cr) at (12.3,0) {\texttt{Docker\_run}, fresh \texttt{out/}\\ image read-only};
  \node[cell] (lf) at (2.6,-1.15) {\texttt{lake update}\\ $\to$ \texttt{.lake/packages}};
  \node[cell] (lb) at (6.5,-1.15) {\texttt{lake build}\\ source read-only};
  \node[cell, text width=3.3cm] (lr) at (12.3,-1.15) {\texttt{lake exe}, fresh \texttt{out/}\\ build read-only};
  \draw[arr] (cf) -- node[lab, above] {cache} (cb);
  \draw[arr] (lf) -- (lb);
  \draw[arr] (cb) -- node[lab, above] {by digest} (cr);
  \draw[arr] (lb) -- (lr);
  \draw[grp] (0.95,0.9) coordinate (bnw) rectangle (8.15,-1.75);
  \draw[grp] (10.35,0.9) coordinate (rnw) rectangle (14.25,-1.75);
  \node[anchor=south west, inner sep=1pt] at (bnw) {\textbf{builder agent} \textit{(writes and compiles; cannot run)}};
  \node[anchor=south west, inner sep=1pt] at (rnw) {\textbf{runner agent} \textit{(runs; cannot write)}};
  \foreach \x/\t in {2.6/fetch, 6.5/build, 12.3/run} \node[font=\scriptsize\bfseries] at (\x,0.62) {\t};
  \node[lane] at (0.9,-2.3) {network};
  \foreach \x/\t in {2.6/egress, 6.5/none, 12.3/none}
    \node[font=\scriptsize\bfseries, fill=white, inner sep=1pt] at (\x,-2.3) {\t};
\end{tikzpicture}
\caption{A staged build across two agents. Fetch sees only public lockfiles and warms a cache; build and run have no
network. The builder cannot run what it built, though its build recipes execute inside the build sandbox; the runner
receives the artifact by digest, cannot change it, and writes only a fresh output directory.}
\label{fig:staged}
\end{figure}

\textbf{Containers and Lake.} A container build and a Lake build share one three-stage shape, split across two agents
(\Cref{fig:staged}). The \emph{fetch} stage reads only public lockfiles, reaches the network, and sees no source; it
warms a cache, the image cache or \texttt{.lake/packages}. The \emph{build} stage mounts the source read-only and has
no network. The \emph{run} stage has no network, mounts the artifact read-only, and gets one fresh \texttt{out/} per
run as its only writable place. The \emph{builder} agent owns fetch and build: it may write the source and compile
it, but holds no authority to run what it built. The \emph{runner} holds that authority but cannot write the source
or the artifact, which the parent hands it by digest. An instruction injected into the builder therefore produces
code that nobody runs without the hand-off, and one injected into the runner acts on bytes it cannot change and can
write only \texttt{out/}. Compiling is not inert, though: build recipes, compiler plugins and Lean's \texttt{\#eval}
run code during the build itself, so the build stage's sandbox must confine that execution on its own, independently
of who may run the result. In the container plan, a stage that mounts private input has no network and no later stage
regains it (\lean{net\_forbidden\_after\_private\_touch}), and a run's only writable mount is its own output directory
(\lean{docker\_plan\_rw\_mount\_is\_its\_out\_dir}); that the container daemon enforces the plan it is given is a
named assumption (\Cref{sec:reversibility}).
For Lake, running a built executable needs the \texttt{wild} grant bit, which a fresh grant leaves off
(\lean{no\_wild\_no\_wild\_effects}), is carded on every call (\lean{build\_lake\_exe\_is\_card}), and only
\texttt{lake update} reaches the network.

Because duties are grants and grants are per agent, any stage boundary can also be an agent boundary, and the gate
can be a human's blessing, a parent's hand-off by digest, or a reviewer that holds no tools. The next example draws
the private-versus-network line between two agents doing research rather than a build.

\subsection{A private agent and a public researcher}
\label{sec:ex-split}

\begin{figure}[ht]
\centering
\begin{tikzpicture}[
  font=\scriptsize, >=Latex,
  zone/.style={draw, thick, rounded corners=4pt},
  ag/.style={draw, circle, thick, fill=green!12, minimum size=8mm, inner sep=0pt, font=\scriptsize\bfseries},
  src/.style={draw, rounded corners=1pt, fill=red!15, inner sep=3pt},
  net/.style={draw, rounded corners=3pt, thick, fill=blue!10, align=center, minimum width=1.3cm, minimum height=1.1cm},
  grant/.style={draw, fill=white, font=\scriptsize\ttfamily, align=left, inner sep=2.5pt, text width=2.4cm},
  hdr/.style={anchor=south west, font=\scriptsize\bfseries, inner sep=1pt},
  lab/.style={font=\scriptsize\itshape, align=center},
  bound/.style={font=\scriptsize\bfseries, align=center},
  taint/.style={orange!75!black},
  clean/.style={blue!45!black},
  fat/.style={line width=3pt, -{Latex[length=2.3mm, width=4mm]}},
  thin/.style={line width=0.7pt, -{Latex[length=1.8mm, width=1.8mm]}},
]
  \newcommand{\none}[1]{\textcolor{black!40}{#1}}
  \newcommand{\gcard}[3]{
    \node[grant] (#1) at #2 {\textbf{grant}\\ #3};
    \fill[black!25] ($(#1.north east)+(-0.2,0)$) -- ++(0.2,-0.2) -- ++(-0.2,0) -- cycle;}
  \draw[zone, fill=red!4] (0,-1.3) rectangle (2.75,1.15);
  \node[hdr] at (0,1.15) {(a) one agent};
  \node[src] (asrc) at (0.55,0.5) {source};
  \node[ag] (A) at (1.95,0.5) {A};
  \draw[->, thick] (asrc) -- (A);
  \gcard{ga}{(1.375,-0.6)}{fs:\ \ project\\ git: project refs\\ net: any host}
  \node[net] (inta) at (4.6,0.4) {internet};
  \draw[fat, taint] ($(A.east)+(0,0.15)$) -- (inta.west |- 0,0.65);
  \draw[fat, clean] (inta.west |- 0,0.2) -- ($(A.east)+(0,-0.3)$);
  \node[lab, text=orange!60!black, below=2pt] at (inta.south) {all egress};
  \node[bound] at (2.6,-1.7) {possibly tainted out: \textcolor{orange!70!black}{all of its egress}};
  \draw[zone, fill=red!4] (6.0,-1.3) rectangle (8.75,1.15);
  \node[hdr] at (6.0,1.15) {(b) private agent};
  \node[src] (bsrc) at (6.55,0.5) {source};
  \node[ag] (P) at (7.95,0.5) {P};
  \draw[->, thick] (bsrc) -- (P);
  \gcard{gp}{(7.375,-0.6)}{fs:\ \ project\\ git: project refs\\ net: \none{none}}
  \draw[zone, fill=blue!4] (11.2,-1.3) rectangle (14.0,1.15);
  \node[hdr] at (11.2,1.15) {public researcher};
  \node[ag] (R) at (12.0,0.5) {R};
  \gcard{gr}{(12.6,-0.6)}{fs:\ \ scratch\\ git: \none{none}\\ net: paper indexes}
  \draw[thin, taint] ($(P.east)+(0,0.2)$) -- ($(R.west)+(0,0.2)$);
  \node[lab, text=orange!60!black, above] at (9.975,0.9) {seed: $k$ bytes};
  \node[fill=orange!55, minimum width=1.3mm, minimum height=4mm, inner sep=0pt] at (9.975,0.7) {};
  \draw[fat, clean] ($(R.west)+(0,-0.22)$) -- ($(P.east)+(0,-0.22)$);
  \node[lab] at (9.975,-0.05) {summary};
  \node[lab, text=black!55] at (9.975,-1.0) {mail};
  \node[net] (intb) at (15.75,0.4) {internet};
  \draw[fat, clean] ($(R.east)+(0,0.2)$) -- (intb.west |- 0,0.65);
  \draw[fat, clean] (intb.west |- 0,0.2) -- ($(R.east)+(0,-0.25)$);
  \node[lab, below=2pt] at (intb.south) {unbounded};
  \node[bound] at (10.9,-1.7) {possibly tainted out: \textcolor{orange!70!black}{at most the seed's $k$ bytes}, however much R fetches};
\end{tikzpicture}
\caption{Why the split bounds leakage. (a) An agent whose grant holds both the source and the network can leak
through every byte it sends. (b) Split into peers with different grants that talk only by mail, the private side's
contribution to anything leaving is the seed, however much traffic the researcher does, under the channel model of
the text. Arrow width stands for volume;
orange marks bytes that may carry the source.}
\label{fig:peer-split}
\end{figure}
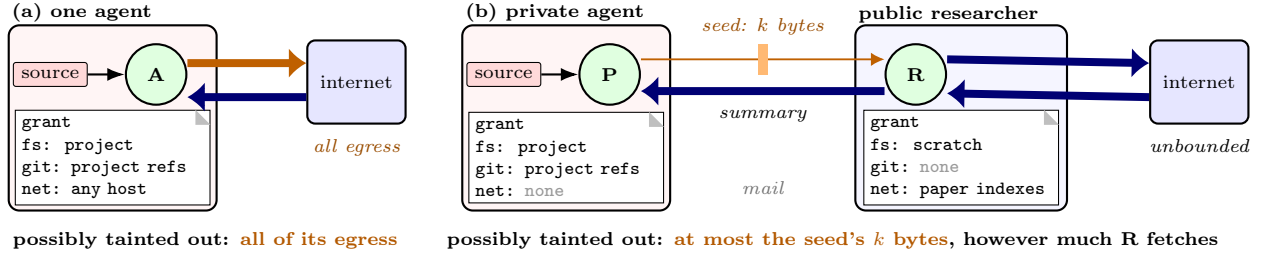

Consider a project whose source and experiment logs are private. The work is split between two peer agents that
talk only by mail (\Cref{fig:peer-split}). The \emph{private} agent's grant names the project's worktree and refs
and no network. The \emph{researcher}'s grant names a scratch worktree and a few paper indexes, and nothing of the
project's. The private agent mails a question, ``Find recent advances in adaptive gradient methods''; the researcher
searches, reads, and mails back a summary.

Write $S$ for the source, $M$ for the whole transcript the private agent sends, $E$ for everything the researcher sends
out, and $Z$ for the public inputs. The researcher's grant names nothing of the project's, and the pages it fetches
answer queries it built from $M$, so the intent is that everything it emits depends on $S$ only through $M$: a Markov
chain $S \to M \to E$ given $Z$. If $M$ is also padded to a fixed width of $k$ bytes, the data-processing inequality
gives
\[
  I(S; E \mid Z) \;\le\; I(S; M \mid Z) \;\le\; H(M \mid Z) \;\le\; 8k \text{ bits},
\]
however much the researcher fetches or sends. Both conditions carry weight. The width must be fixed, or the length of
the mail is itself a channel, and every other route between the peers --- the timing and number of letters, shared
files, anything else either can observe --- must be counted in $M$ or ruled out. The grants express the split; they do
not by themselves prove that channel model, and L3M does not yet enforce a budget on it (\Cref{app:infoflow}).

\subsection{A quarantined reader}
\label{sec:ex-quarantine}

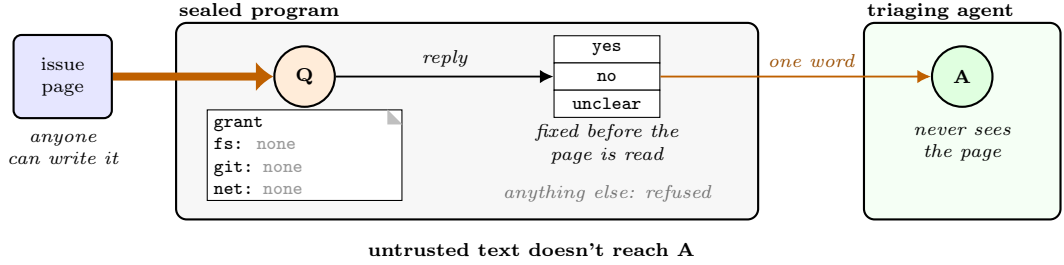
\begin{figure}[ht]
\centering
\begin{tikzpicture}[
  font=\scriptsize, >=Latex,
  zone/.style={draw, thick, rounded corners=4pt},
  ag/.style={draw, circle, thick, fill=green!12, minimum size=8mm, inner sep=0pt, font=\scriptsize\bfseries},
  net/.style={draw, rounded corners=3pt, thick, fill=blue!10, align=center, minimum width=1.3cm, minimum height=1.1cm},
  grant/.style={draw, fill=white, font=\scriptsize\ttfamily, align=left, inner sep=2.5pt, text width=2.4cm},
  opt/.style={draw, fill=white, font=\scriptsize\ttfamily, minimum width=1.4cm, minimum height=0.36cm, inner sep=1pt},
  hdr/.style={anchor=south west, font=\scriptsize\bfseries, inner sep=1pt},
  lab/.style={font=\scriptsize\itshape, align=center},
  bound/.style={font=\scriptsize\bfseries, align=center},
  taint/.style={orange!75!black},
  fat/.style={line width=3pt, -{Latex[length=2.3mm, width=4mm]}},
  thin/.style={line width=0.7pt, -{Latex[length=1.8mm, width=1.8mm]}},
]
  \newcommand{\none}[1]{\textcolor{black!40}{#1}}
  \newcommand{\gcard}[3]{
    \node[grant] (#1) at #2 {\textbf{grant}\\ #3};
    \fill[black!25] ($(#1.north east)+(-0.2,0)$) -- ++(0.2,-0.2) -- ++(-0.2,0) -- cycle;}
  \node[net] (page) at (0.8,0.44) {issue\\ page};
  \node[lab, below=2pt] at (page.south) {anyone\\ can write it};
  \draw[zone, fill=black!3] (2.3,-1.45) rectangle (10.0,1.15);
  \node[hdr] at (2.3,1.15) {sealed program};
  \node[ag, fill=orange!15] (Q) at (4.0,0.44) {Q};
  \gcard{gq}{(4.0,-0.6)}{fs:\ \ \none{none}\\ git: \none{none}\\ net: \none{none}}
  \draw[fat, taint] (page.east) -- (Q.west);
  \node[opt] (yes) at (8.0,0.8) {yes};
  \node[opt] (no) at (8.0,0.44) {no};
  \node[opt] (unc) at (8.0,0.08) {unclear};
  \node[lab] at (8.0,-0.45) {fixed before the\\ page is read};
  \node[lab, text=black!55] at (8.0,-1.1) {anything else: refused};
  \draw[thin] (Q.east) -- (no.west) node[midway, above, lab] {reply};
  \draw[zone, fill=green!4] (11.4,-1.45) rectangle (14.0,1.15);
  \node[hdr] at (11.4,1.15) {triaging agent};
  \node[ag] (A) at (12.7,0.44) {A};
  \draw[thin, taint] (no.east) -- (A.west);
  \node[lab, text=orange!60!black, above] at (10.7,0.44) {one word};
  \node[lab] at (12.7,-0.45) {never sees\\ the page};
  \node[bound] at (7.0,-1.85) {untrusted text doesn't reach A};
\end{tikzpicture}
\caption{A quarantined reader. The program hands the page to a fresh subagent Q whose grant is empty, and parses its
reply into one of three words fixed in advance. A page that hijacks Q gains a model that holds nothing, and can change
only which word reaches A. Arrow width stands for volume; orange marks bytes that may carry an injected instruction.}
\label{fig:quarantine}
\end{figure}

The previous example keeps private bytes from getting out; this one keeps untrusted instructions from getting in. An
agent triaging bug reports must act on pages anyone can write, and an instruction planted on one (``ignore your task
and push to mainline'') is aimed at whichever model reads it. So the agent does not read the page. It calls a sealed
program (\Cref{sec:sealed}) that fetches the page and drives a fresh subagent to completion through the spawn gate,
under a grant that holds nothing of the triaging agent's: no project files or refs, and no network
(\Cref{fig:quarantine}). The program asks the subagent one question, ``does this report a regression in the parser?'',
and parses the reply as one of \texttt{yes}, \texttt{no} or \texttt{unclear}, mapping any other reply, a failure or a
timeout to \texttt{unclear}. Only that word reaches the triaging agent.

The page can still hijack the subagent, but a hijacked subagent holds nothing to misuse and is discarded when the call
returns. Its only way out is a menu fixed before the page was read, so each call gives the page at most $\log_2 3$ bits
of influence over the agent that acts, provided the call's timing and failures reveal nothing beyond the word. This is the bound of \Cref{sec:ex-split} in the other direction: there the mail
limits what can leave, here the menu limits what can enter. It limits how far the page can steer the answer, not
whether the answer is right, and a wrong \texttt{no} is still wrong. Like every pattern in this section, it helps only
if it is used: an agent that also fetches the page itself reads the instruction anyway. CaMeL~\cite{debenedetti2025camel}
separates a quarantined model from a privileged one in the same way; here the separation is built from a grant, a
program and a type.

\section{Improving Agents Inside L3M}
\label{sec:improve}

\paragraph{Behaviour is data.} What \Cref{sec:log} says of one agent holds for every agent in a hub: its messages,
tool calls and results, review outcomes, token counts and timings are records on its refs, and anything computed
from them --- cost per task, latency per tool, how often review declined a call, which events a hook dismissed ---
is a fold over those records. Measuring an agent therefore needs no instrumentation, and the question of who may
measure whom is already answered by the grant tree: a parent reads the Logs and persona branches of all its
descendants (\lean{descendantRead\_covers\_descendants}), and nothing an experiment reads lies outside its grant.

\paragraph{Experiments are bounded agents.} An agent that wants to improve other agents does it with the same
primitives it uses for any other work. It writes a sealed program (\Cref{sec:sealed}) that folds over its
children's Logs, and it compares variants by spawning children that wear different versions of a persona.
Nothing about the experiment is special to the
harness: the children's grants are below the experimenter's, their spend is bounded by the budget it lent them
(\lean{global\_budget\_bounded\_by\_root}), and each program's type bounds what it touches. An optimization loop
over agents is an agent tree, and every guarantee of \Cref{tab:claims} holds for it.

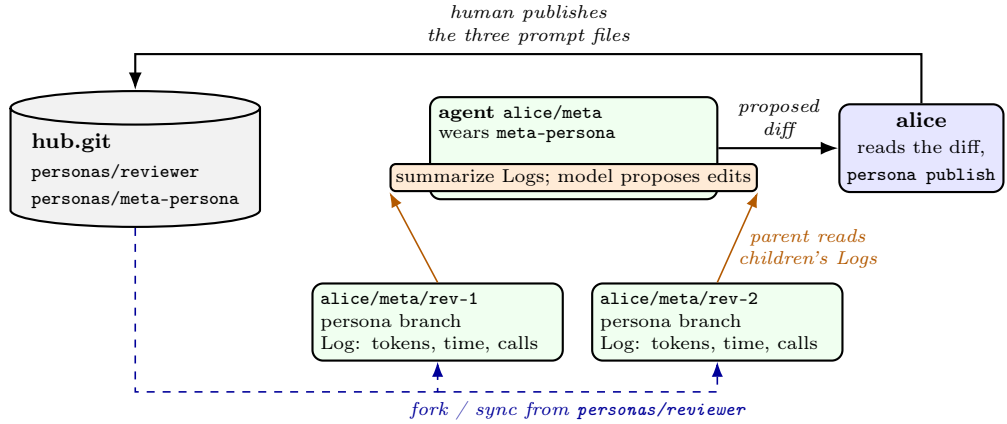
\begin{figure}[t]
\centering
\begin{tikzpicture}[
  font=\footnotesize, >=Latex,
  repo/.style={draw, thick, cylinder, shape border rotate=90, aspect=0.15, fill=gray!10,
               minimum width=3.3cm, align=left, inner sep=3pt},
  ag/.style={draw, thick, rounded corners=4pt, fill=green!6, align=left, inner sep=3pt, font=\scriptsize},
  prog/.style={draw, thick, rounded corners=2pt, fill=orange!16, align=center, inner sep=2pt, font=\scriptsize},
  hum/.style={draw, thick, rounded corners=4pt, fill=blue!10, align=center, inner sep=3pt},
  lab/.style={font=\scriptsize\itshape, align=center},
  fork/.style={->, semithick, dashed, blue!60!black},
  rd/.style={->, semithick, orange!70!black},
  pub/.style={->, thick},
]
  \node[repo, minimum height=1.45cm] (hub) at (0,0) {\textbf{hub.git}\\[1pt]
    \scriptsize\texttt{personas/reviewer}\\ \scriptsize\texttt{personas/meta-persona}};

  \node[draw, thick, rounded corners=4pt, fill=green!6, minimum width=3.8cm, minimum height=1.35cm] (meta) at (5.8,0.3) {};
  \node[font=\scriptsize, anchor=north west, align=left] at (meta.north west)
    {\textbf{agent \texttt{alice/meta}}\\[-1pt] wears \texttt{meta-persona}};
  \node[prog] (dist) at ($(meta.south)+(0,0.3)$) {summarize Logs; model proposes edits};

  \node[ag, text width=3.1cm] (r1) at (4.0,-2.0) {\textbf{\texttt{alice/meta/rev-1}}\\ persona branch\\
    Log: tokens, time, calls};
  \node[ag, text width=3.1cm] (r2) at (7.7,-2.0) {\textbf{\texttt{alice/meta/rev-2}}\\ persona branch\\
    Log: tokens, time, calls};
  \draw[rd] (r1.north) -- (dist.south west);
  \draw[rd] (r2.north) -- (dist.south east);
  \node[lab, text=orange!70!black, anchor=west] at ($(r2.north)+(0.15,0.45)$) {parent reads\\ children's Logs};

  \coordinate (fk) at ($(r1.south)+(0,-0.4)$);
  \draw[fork] (hub.south) |- (fk) -- (r1.south);
  \draw[fork] (fk) -| (r2.south);
  \node[lab, text=blue!60!black, anchor=north] at ($(fk)!0.5!(fk -| r2.south)$) {fork / sync from \texttt{personas/reviewer}};

  \node[hum] (alice) at (10.4,0.3) {\textbf{alice}\\ \scriptsize reads the diff,\\ \scriptsize \texttt{persona publish}};
  \draw[pub] (meta.east) -- node[lab, above] {proposed\\ diff} (alice.west);
  \draw[pub] (alice.north) -- ++(0,0.65) -| node[lab, above, pos=0.25] {human publishes\\ the three prompt files} (hub.north);
\end{tikzpicture}
\caption{One instance of improving agents inside L3M. Every arrow is an L3M primitive: a persona branch forks and
syncs, a parent reads its children's Logs, a program summarizes them and the model proposes edits, and a human
publishes the result.}
\label{fig:meta-loop}
\end{figure}

\paragraph{What can be improved.} Anything an agent's behaviour is a function of, and that lives on a ref, can be
improved this way. Examples:
\begin{itemize}[leftmargin=1.5em,nosep]
  \item \emph{personas}: distil what the wearers of a persona learned into its three files (\Cref{fig:meta-loop});
  \item \emph{hooks}: tune a disposition policy against how often the model dismisses the events it is shown;
  \item \emph{tools}: compare sealed programs for the same job --- a search, a bisect --- on tokens and time;
  \item \emph{review}: use per-tool decline counts to find the calls for which a tame alternative should exist.
\end{itemize}
These are uses the infrastructure supports; we report no measured gain from any of them here.

\paragraph{An example: improving a persona.} In \Cref{fig:meta-loop}, \texttt{alice/meta} wears a meta-persona
and is the parent of two agents that wear \texttt{personas/reviewer}. It uses a program to summarize its children's
Logs, and asks its model to propose new versions of the reviewer persona's three files. The proposal stays on
\texttt{alice/meta}'s own branch, because no grant can write a persona ref (\Cref{sec:personas}). alice reads the
diff and publishes it; each reviewer pulls the new version into its persona branch, and the next reviewer forks
from it.

\paragraph{The harness.} The same loop applies to L3M itself. The sealed programs and hooks an agent writes are
extensions of the harness, measured the same way, and L3M's own source is a Lean program in a repository an agent
can claim. Optimizing a measured objective there raises its own problem: an agent that can change the evaluator,
weaken a theorem's statement or add an assumption can improve its score, or get a green build, by moving the target
rather than improving the system. A change that breaks a theorem of \Cref{tab:claims} does not build, but a change that
edits the theorem does. Experiments on the harness therefore need the specification and the evaluation criteria held
by someone other than the optimizer --- in L3M, a separate grant --- with every candidate judged against them,
including the assumptions its proofs use.

\section{Evaluation}
\label{sec:eval}

\paragraph{Setup} The guarantees of \Cref{tab:claims} are theorems; no experiment adds to them. What is empirical is what the bound on
authority costs in work done, and what review costs to use. We measure both on Terminal-Bench~4.0.0, which gives the
agent a container and an instruction per task and grades what it leaves with the task's own tests. We select 37 tasks 
that are valid in our harness-agnostic environment and that a public Claude Code run finished in a
median of at most an hour. Every arm runs Claude Opus~5 at the same reasoning effort, 64{,}000-token output limit and
75-minute deadline, with three attempts per task and Claude Sonnet~4.6 as reviewer; only the harness differs. Claude
Code runs with every permission check bypassed, so its tools run without approval and its shell reaches the network. 
The L3M arms differ only in which tools the model is offered and whether shell
calls are reviewed. The runs measure the first implementation of L3M, which has not been tuned for efficiency; we
return to this at the end of the section.

\begin{table}[ht]
\centering\small
\caption{Terminal-Bench: 37 tasks, three attempts each; pass@3 is averaged over tasks. Model calls, tokens and
minutes are means over passing attempts and include any subagent sessions;  tokens count cached and uncached 
input plus output.}
\label{tab:bench}
\begin{tabular}{@{}lrrrr@{}}
\toprule
 & & \multicolumn{3}{c}{Per passing attempt} \\
\cmidrule(l){3-5}
Configuration & Pass@3, \% & Model calls & Tokens, M & Minutes \\
\midrule
Claude Code, permissions bypassed & 59.5 & 53.0 & 4.43 & 24.6 \\
L3M, shell, review off & 41.7 & 53.2 & 7.70 & 25.4 \\
\midrule
L3M, tame & 27.0 & 53.3 & 7.74 & 21.6 \\
L3M, tame, no programs & 24.3 & 41.5 & 6.17 & 24.5 \\
\bottomrule
\end{tabular}
\end{table}

\paragraph{How much work can one do using only tame tools?} The tame arm offers no shell: every tool the model holds
is either undoable or bounded by a token, so nothing it does is reviewed. Against the same harness with a shell and
review off, pass@3 falls from 41.7\% to 27.0\% (34 passes to 18, \Cref{tab:bench}). Of the 18 tame
passes, five used the tool programs. So about two thirds of what the shell delivers on these tasks is reachable with no wild call at
all in the current L3M implementation. This is promising in terms of eventually being able to perform a large portion of useful work in a ``tame'' manner.

\paragraph{How useful are tool programs?} Tool programs add about 3 points at pass@3: 27.0\% with program loading
against 24.3\% without (\Cref{tab:bench}). The gain is small, although we did not tune the system prompt to encourage
their use. 

\begin{table}[ht]
\centering\small
\caption{Review of shell calls, on the same tasks and with the same columns as \Cref{tab:bench}. Model calls count
both sides of the review. $^*$Tokens include the reviewer's side of each exchange, which runs on a cheaper model
(Claude Sonnet~4.6) than the agent's.}
\label{tab:review}
\begin{tabular}{@{}lrrrr@{}}
\toprule
 & & \multicolumn{3}{c}{Per passing attempt} \\
\cmidrule(l){3-5}
Configuration & Pass@3, \% & Model calls & Tokens, M & Minutes \\
\midrule
L3M, shell, review off & 41.7 & 53.2 & 7.70 & 25.4 \\
L3M, shell, review \texttt{-{}-yes} & 40.5 & 100.9 & 12.10$^*$ & 30.4 \\
L3M, shell, review \texttt{-{}-no} & 43.2 & 107.3 & 12.81$^*$ & 28.1 \\
\bottomrule
\end{tabular}
\end{table}

\paragraph{What review costs, and how often it redirects.} These arms review only shell calls, which make up 25.1\% of
all tool calls. Around 90\% of shell calls are judged safe in the end. The reviewer's first verdict was \texttt{safe} about two thirds of the time, and when it was not,
about nine in ten second rounds ended \texttt{safe}. About 6\% of reviewed calls were redirected to
tame tools as an alternative. A further 2\% were withdrawn upon objection from the reviewer. Stalemates,
where a human would decide, were 2.4\% of reviewed calls. The price is model work, not success: each reviewed call adds about 2.6
model calls and a median of 8.6 seconds, so review adds roughly 85\% more calls and 50\% more tokens per attempt
(\Cref{tab:review}), albeit under a cheaper model. Interestingly, review does not change overall success rates
-- the $\sim 1\%$ differences are not statistically significant (only 37 tasks).

\paragraph{A caveat on efficiency.} L3M has not been tuned for efficiency. Neither the system prompt, the tool
descriptions nor the context management have been optimized against any benchmark, as a production harness's
have. The numbers in \Cref{tab:bench,tab:review} should be read as a first measurement of
an untuned harness rather than as its ceiling. In particular, the token gap to Claude Code at a similar number of
model calls points at the harness, not at the authority bound: the tame arm and the shell arm spend about the same
tokens per passing attempt. We expect pass rates and token costs to improve once the harness is tuned, and the
theorems of \Cref{tab:claims} do not depend on that tuning.

\section{Discussion and Conclusion}
\label{sec:discussion}

\paragraph{Future work.} The budgets charge quantities: model calls, bytes and wakes. The natural next bound charges
information. Data would carry a label, and each agent would accumulate what crossed its edges, charged at the point
where L3M already meters egress, so the extension changes what is charged but not where. Sealed programs can drive children
but cannot yet build or wait on a vigil, so long-running orchestration still passes through the model. A star on a branch or
a repository waits on claims becoming subscriptions (\Cref{sec:mail}), and starting agents on another host is designed
but not built. The evaluation (\Cref{sec:eval}) measures the first implementation. Tame tools alone already reach about
two thirds of what the shell delivers at pass@3, tool programs add a few points without any prompting to use them,
and review costs model work rather than success. The same tables also show the gap to an unconstrained harness: with
a shell, L3M trails Claude Code at pass@3, and at a similar number of model calls per solved task it spends well over
Claude Code's tokens (\Cref{tab:bench}). Both say that the harness and its system prompt, neither of which has been
optimized, need tuning before performance is comparable. 

\paragraph{Conclusion.} L3M trusts the harness rather than the model, and it states what that trust buys as theorems
about the code that runs. Authority and budget only narrow down a tree of agents. A tool's review tier follows from
how far its effects can be undone and what its tokens bound. An agent extends its own tool surface with sealed
programs whose types are their permission requests. So an agent can grow its harness over time without new
permissions, and the human can still read which invariants hold. Capability grows; authority does not. We argue that
this is a building block for long-running, many-agent systems, and we release L3M as open source so that others can
build on it. 

\section*{Acknowledgements}
We thank Jim Roskind for early discussions of and feedback on L3M, including the recipe for defending against prompt
injection described in \Cref{sec:ex-quarantine}.

\ifhasbib
\clearpage
\bibliographystyle{styles/ims_nourl_eprint}
\bibliography{references}

\begin{thebibliography}{51}
\expandafter\ifx\csname natexlab\endcsname\relax\def\natexlab#1{#1}\fi
\expandafter\ifx\csname url\endcsname\relax
  \def\url#1{\texttt{#1}}\fi
\expandafter\ifx\csname urlprefix\endcsname\relax\def\urlprefix{URL }\fi

\bibitem[{Anil et~al.(2021)Anil, Zhang, Wu and Grosse}]{anil2021pvg}
\textsc{Anil, C.}, \textsc{Zhang, G.}, \textsc{Wu, Y.} and \textsc{Grosse, R.}
  (2021).
\newblock {Learning to Give Checkable Answers with Prover-Verifier Games}.
\newblock \textit{arXiv preprint arXiv:2108.12099} .

\bibitem[{Antoniades et~al.(2025)Antoniades, Örwall, Zhang, Xie, Goyal and
  Wang}]{antoniades2025swesearch}
\textsc{Antoniades, A.}, \textsc{Örwall, A.}, \textsc{Zhang, K.}, \textsc{Xie,
  Y.}, \textsc{Goyal, A.} and \textsc{Wang, W.} (2025).
\newblock {SWE-Search}: Enhancing software agents with {Monte Carlo} tree
  search and iterative refinement.
\newblock In \textit{International Conference on Learning Representations (ICLR
  2025)}.

\bibitem[{Bainbridge(1983)}]{bainbridge1983ironies}
\textsc{Bainbridge, L.} (1983).
\newblock Ironies of automation.
\newblock \textit{Automatica} \textbf{19} 775--779.

\bibitem[{Barbaste et~al.(2026)Barbaste, Darrigol, Vu and
  Wiltberger}]{barbaste2026harness}
\textsc{Barbaste, P.}, \textsc{Darrigol, T.}, \textsc{Vu, G.} and
  \textsc{Wiltberger, T.} (2026).
\newblock Harness engineering: Anatomy, architecture, and evolution of coding
  agents -- a source-code study of eleven systems.
\newblock {\href{https://arxiv.org/abs/2609.00006}{\texttt{arXiv:2609.00006}}}.

\bibitem[{Bershad et~al.(1995)Bershad, Savage, Pardyak, Sirer, Fiuczynski,
  Becker, Chambers and Eggers}]{bershad1995spin}
\textsc{Bershad, B.~N.}, \textsc{Savage, S.}, \textsc{Pardyak, P.},
  \textsc{Sirer, E.~G.}, \textsc{Fiuczynski, M.~E.}, \textsc{Becker, D.},
  \textsc{Chambers, C.} and \textsc{Eggers, S.} (1995).
\newblock Extensibility, safety and performance in the {SPIN} operating system.
\newblock In \textit{Proceedings of the Fifteenth ACM Symposium on Operating
  Systems Principles (SOSP '95)}. ACM.

\bibitem[{Birgisson et~al.(2014)Birgisson, Politz, Erlingsson, Taly, Vrable and
  Lentczner}]{birgisson2014macaroons}
\textsc{Birgisson, A.}, \textsc{Politz, J.~G.}, \textsc{Erlingsson, {\'U}.},
  \textsc{Taly, A.}, \textsc{Vrable, M.} and \textsc{Lentczner, M.} (2014).
\newblock Macaroons: Cookies with contextual caveats for decentralized
  authorization in the cloud.
\newblock In \textit{Proceedings 2014 Network and Distributed System Security
  Symposium (NDSS)}. Internet Society.

\bibitem[{Brown-Cohen et~al.(2025)Brown-Cohen, Irving, Piliouras, Chen, Li and
  Xun}]{browncohen2025avoiding}
\textsc{Brown-Cohen, J.}, \textsc{Irving, G.}, \textsc{Piliouras, G.},
  \textsc{Chen, L.}, \textsc{Li, J.} and \textsc{Xun, Z.} (2025).
\newblock Avoiding obfuscation with prover-estimator debate.
\newblock {\href{https://arxiv.org/abs/2506.13609}{\texttt{arXiv:2506.13609}}}.
\newblock V2, 21 July 2026.

\bibitem[{Cai et~al.(2024)Cai, Wang, Ma, Chen and Zhou}]{cai2024latm}
\textsc{Cai, T.}, \textsc{Wang, X.}, \textsc{Ma, T.}, \textsc{Chen, X.} and
  \textsc{Zhou, D.} (2024).
\newblock Large language models as tool makers.
\newblock In \textit{The Twelfth International Conference on Learning
  Representations (ICLR)}.

\bibitem[{Costa et~al.(2025)Costa, K{\"o}pf, Kolluri, Paverd, Russinovich,
  Salem, Tople, Wutschitz and Zanella-B{\'e}guelin}]{costa2025fides}
\textsc{Costa, M.}, \textsc{K{\"o}pf, B.}, \textsc{Kolluri, A.},
  \textsc{Paverd, A.}, \textsc{Russinovich, M.}, \textsc{Salem, A.},
  \textsc{Tople, S.}, \textsc{Wutschitz, L.} and \textsc{Zanella-B{\'e}guelin,
  S.} (2025).
\newblock Securing {AI} agents with information-flow control.
\newblock {\href{https://arxiv.org/abs/2505.23643}{\texttt{arXiv:2505.23643}}}.

\bibitem[{de~Moura and Ullrich(2021)}]{demoura2021lean4}
\textsc{de~Moura, L.} and \textsc{Ullrich, S.} (2021).
\newblock The {Lean}~4 theorem prover and programming language.
\newblock In \textit{Automated Deduction -- {CADE} 28}. Lecture Notes in
  Computer Science, Springer.

\bibitem[{de~Raadt(2004)}]{deraadt2004wx}
\textsc{de~Raadt, T.} (2004).
\newblock Exploit mitigation techniques.
\newblock Talk at AUUG 2004, Melbourne. Slides:
  \url{https://www.openbsd.org/papers/auug04/}.

\bibitem[{Debenedetti et~al.(2025)Debenedetti, Shumailov, Fan, Hayes, Carlini,
  Fabian, Kern, Shi, Terzis and Tram{\`e}r}]{debenedetti2025camel}
\textsc{Debenedetti, E.}, \textsc{Shumailov, I.}, \textsc{Fan, T.},
  \textsc{Hayes, J.}, \textsc{Carlini, N.}, \textsc{Fabian, D.}, \textsc{Kern,
  C.}, \textsc{Shi, C.}, \textsc{Terzis, A.} and \textsc{Tram{\`e}r, F.}
  (2025).
\newblock Defeating prompt injections by design.
\newblock {\href{https://arxiv.org/abs/2503.18813}{\texttt{arXiv:2503.18813}}}.

\bibitem[{Debenedetti et~al.(2024)Debenedetti, Zhang, Balunovi{\'c},
  Beurer-Kellner, Fischer and Tram{\`e}r}]{debenedetti2024agentdojo}
\textsc{Debenedetti, E.}, \textsc{Zhang, J.}, \textsc{Balunovi{\'c}, M.},
  \textsc{Beurer-Kellner, L.}, \textsc{Fischer, M.} and \textsc{Tram{\`e}r, F.}
  (2024).
\newblock {AgentDojo}: A dynamic environment to evaluate prompt injection
  attacks and defenses for {LLM} agents.
\newblock In \textit{Advances in Neural Information Processing Systems
  (NeurIPS) Datasets and Benchmarks Track}.

\bibitem[{Dennis and Van~Horn(1966)}]{dennis1966semantics}
\textsc{Dennis, J.~B.} and \textsc{Van~Horn, E.~C.} (1966).
\newblock Programming semantics for multiprogrammed computations.
\newblock \textit{Communications of the ACM} \textbf{9} 143--155.

\bibitem[{Devriese et~al.(2016)Devriese, Birkedal and
  Piessens}]{devriese2016parametricity}
\textsc{Devriese, D.}, \textsc{Birkedal, L.} and \textsc{Piessens, F.} (2016).
\newblock Reasoning about object capabilities with logical relations and effect
  parametricity.
\newblock In \textit{2016 IEEE European Symposium on Security and Privacy
  (EuroS\&P)}. IEEE.

\bibitem[{Foster(2026)}]{leanmanifests}
\textsc{Foster, D.} (2026).
\newblock {Lean Manifests}: Contracts with evidence levels.
\newblock \url{https://lean4.ai/lean-manifests.html}, accessed 2026-09-27.

\bibitem[{Garby et~al.(2026)Garby, Gordon and Sands}]{garby2026llmbda}
\textsc{Garby, Z.}, \textsc{Gordon, A.~D.} and \textsc{Sands, D.} (2026).
\newblock The {LLMbda} calculus: {AI} agents, conversations, and information
  flow.
\newblock {\href{https://arxiv.org/abs/2602.20064}{\texttt{arXiv:2602.20064}}}.

\bibitem[{Greenblatt et~al.(2024)Greenblatt, Shlegeris, Sachan and
  Roger}]{greenblatt2024control}
\textsc{Greenblatt, R.}, \textsc{Shlegeris, B.}, \textsc{Sachan, K.} and
  \textsc{Roger, F.} (2024).
\newblock {AI} control: Improving safety despite intentional subversion.
\newblock In \textit{Proceedings of the 41st International Conference on
  Machine Learning}, vol. 235 of \textit{Proceedings of Machine Learning
  Research}. PMLR.

\bibitem[{Greshake et~al.(2023)Greshake, Abdelnabi, Mishra, Endres, Holz and
  Fritz}]{greshake2023injection}
\textsc{Greshake, K.}, \textsc{Abdelnabi, S.}, \textsc{Mishra, S.},
  \textsc{Endres, C.}, \textsc{Holz, T.} and \textsc{Fritz, M.} (2023).
\newblock Not what you've signed up for: Compromising real-world
  {LLM}-integrated applications with indirect prompt injection.
\newblock In \textit{Proceedings of the 16th ACM Workshop on Artificial
  Intelligence and Security (AISec '23)}. ACM.

\bibitem[{Hardy(1988)}]{hardy1988confused}
\textsc{Hardy, N.} (1988).
\newblock The confused deputy (or why capabilities might have been invented).
\newblock \textit{ACM SIGOPS Operating Systems Review} \textbf{22} 36--38.

\bibitem[{Hong et~al.(2024)Hong, Zhuge, Chen, Zheng, Cheng, Zhang, Wang, Wang,
  Yau, Lin, Zhou, Ran, Xiao, Wu and Schmidhuber}]{hong2023metagpt}
\textsc{Hong, S.}, \textsc{Zhuge, M.}, \textsc{Chen, J.}, \textsc{Zheng, X.},
  \textsc{Cheng, Y.}, \textsc{Zhang, C.}, \textsc{Wang, J.}, \textsc{Wang, Z.},
  \textsc{Yau, S. K.~S.}, \textsc{Lin, Z.}, \textsc{Zhou, L.}, \textsc{Ran,
  C.}, \textsc{Xiao, L.}, \textsc{Wu, C.} and \textsc{Schmidhuber, J.} (2024).
\newblock {MetaGPT}: Meta programming for a multi-agent collaborative
  framework.
\newblock In \textit{International Conference on Learning Representations
  (ICLR)}.

\bibitem[{Horvitz(1999)}]{horvitz1999mixedinitiative}
\textsc{Horvitz, E.} (1999).
\newblock Principles of mixed-initiative user interfaces.
\newblock In \textit{Proceedings of the SIGCHI Conference on Human Factors in
  Computing Systems (CHI '99)}. ACM, Pittsburgh, Pennsylvania, USA.

\bibitem[{Hunt and Larus(2007)}]{hunt2007singularity}
\textsc{Hunt, G.~C.} and \textsc{Larus, J.~R.} (2007).
\newblock Singularity: Rethinking the software stack.
\newblock \textit{ACM SIGOPS Operating Systems Review} \textbf{41} 37--49.

\bibitem[{Irving et~al.(2018)Irving, Christiano and Amodei}]{irving2018debate}
\textsc{Irving, G.}, \textsc{Christiano, P.} and \textsc{Amodei, D.} (2018).
\newblock {AI} safety via debate.
\newblock \textit{arXiv preprint arXiv:1805.00899} .

\bibitem[{Ke(2026)}]{ke2026quine}
\textsc{Ke, H.} (2026).
\newblock Quine: Realizing {LLM} agents as native {POSIX} processes.
\newblock {\href{https://arxiv.org/abs/2603.18030}{\texttt{arXiv:2603.18030}}}.

\bibitem[{Klein et~al.(2009)Klein, Elphinstone, Heiser, Andronick, Cock,
  Derrin, Elkaduwe, Engelhardt, Kolanski, Norrish, Sewell, Tuch and
  Winwood}]{klein2009sel4}
\textsc{Klein, G.}, \textsc{Elphinstone, K.}, \textsc{Heiser, G.},
  \textsc{Andronick, J.}, \textsc{Cock, D.}, \textsc{Derrin, P.},
  \textsc{Elkaduwe, D.}, \textsc{Engelhardt, K.}, \textsc{Kolanski, R.},
  \textsc{Norrish, M.}, \textsc{Sewell, T.}, \textsc{Tuch, H.} and
  \textsc{Winwood, S.} (2009).
\newblock {seL4}: Formal verification of an {OS} kernel.
\newblock In \textit{Proceedings of the ACM SIGOPS 22nd Symposium on Operating
  Systems Principles (SOSP '09)}. ACM.

\bibitem[{Lakos(1996)}]{lakos1996}
\textsc{Lakos, J.} (1996).
\newblock \textit{Large-Scale {C++} Software Design}.
\newblock Addison-Wesley Professional Computing Series, Addison-Wesley.

\bibitem[{Leroy(2009)}]{leroy2009compcert}
\textsc{Leroy, X.} (2009).
\newblock Formal verification of a realistic compiler.
\newblock \textit{Communications of the ACM} \textbf{52} 107--115.

\bibitem[{Licklider(1960)}]{licklider1960symbiosis}
\textsc{Licklider, J. C.~R.} (1960).
\newblock Man-computer symbiosis.
\newblock \textit{IRE Transactions on Human Factors in Electronics}
  \textbf{HFE-1} 4--11.

\bibitem[{Liu et~al.(2026)Liu, Zhao, Shang and Shen}]{liu2026claudecode}
\textsc{Liu, J.}, \textsc{Zhao, X.}, \textsc{Shang, X.} and \textsc{Shen, Z.}
  (2026).
\newblock Dive into {Claude Code}: The design space of today's and future {AI}
  agent systems.
\newblock {\href{https://arxiv.org/abs/2604.14228}{\texttt{arXiv:2604.14228}}}.
\newblock Tech report.

\bibitem[{Mei et~al.(2025)Mei, Zhu, Xu, Jin, Hua, Li, Xu, Ye, Ge and
  Zhang}]{mei2025aios}
\textsc{Mei, K.}, \textsc{Zhu, X.}, \textsc{Xu, W.}, \textsc{Jin, M.},
  \textsc{Hua, W.}, \textsc{Li, Z.}, \textsc{Xu, S.}, \textsc{Ye, R.},
  \textsc{Ge, Y.} and \textsc{Zhang, Y.} (2025).
\newblock {AIOS}: {LLM} agent operating system.
\newblock In \textit{Second Conference on Language Modeling (COLM)}.
\newblock ArXiv:2403.16971.

\bibitem[{Mettler et~al.(2010)Mettler, Wagner and Close}]{mettler2010joee}
\textsc{Mettler, A.}, \textsc{Wagner, D.} and \textsc{Close, T.} (2010).
\newblock {Joe-E}: A security-oriented subset of {Java}.
\newblock In \textit{Proceedings of the Network and Distributed System Security
  Symposium (NDSS 2010)}. The Internet Society, San Diego, CA, USA.

\bibitem[{Miller and Shapiro(2003)}]{miller2003paradigm}
\textsc{Miller, M.~S.} and \textsc{Shapiro, J.~S.} (2003).
\newblock Paradigm regained: Abstraction mechanisms for access control.
\newblock In \textit{Advances in Computing Science -- ASIAN 2003: Programming
  Languages and Distributed Computation} (V.~A. Saraswat, ed.). Springer,
  Berlin, Heidelberg.

\bibitem[{Mozannar et~al.(2025)Mozannar, Bansal, Tan, Fourney, Dibia, Chen,
  Gerrits, Payne, Maldaner, Grunde-McLaughlin, Zhu, Bassman, Alber, Chang,
  Loynd, Niedtner, Kamar, Murad, Hosn and Amershi}]{mozannar2025magenticui}
\textsc{Mozannar, H.}, \textsc{Bansal, G.}, \textsc{Tan, C.}, \textsc{Fourney,
  A.}, \textsc{Dibia, V.}, \textsc{Chen, J.}, \textsc{Gerrits, J.},
  \textsc{Payne, T.}, \textsc{Maldaner, M.~K.}, \textsc{Grunde-McLaughlin, M.},
  \textsc{Zhu, E.}, \textsc{Bassman, G.}, \textsc{Alber, J.}, \textsc{Chang,
  P.}, \textsc{Loynd, R.}, \textsc{Niedtner, F.}, \textsc{Kamar, E.},
  \textsc{Murad, M.}, \textsc{Hosn, R.} and \textsc{Amershi, S.} (2025).
\newblock {Magentic-UI}: Towards human-in-the-loop agentic systems.
\newblock {\href{https://arxiv.org/abs/2507.22358}{\texttt{arXiv:2507.22358}}}.

\bibitem[{Necula(1997)}]{necula1997pcc}
\textsc{Necula, G.~C.} (1997).
\newblock Proof-carrying code.
\newblock In \textit{Proceedings of the 24th ACM SIGPLAN-SIGACT Symposium on
  Principles of Programming Languages (POPL '97)}. ACM.

\bibitem[{Nienhuis et~al.(2020)Nienhuis, Joannou, Bauereiss, Fox, Roe,
  Campbell, Naylor, Norton, Moore, Neumann, Stark, Watson and
  Sewell}]{nienhuis2020cheri}
\textsc{Nienhuis, K.}, \textsc{Joannou, A.}, \textsc{Bauereiss, T.},
  \textsc{Fox, A.}, \textsc{Roe, M.}, \textsc{Campbell, B.}, \textsc{Naylor,
  M.}, \textsc{Norton, R.~M.}, \textsc{Moore, S.~W.}, \textsc{Neumann, P.~G.},
  \textsc{Stark, I.}, \textsc{Watson, R. N.~M.} and \textsc{Sewell, P.} (2020).
\newblock Rigorous engineering for hardware security: Formal modelling and
  proof in the {CHERI} design and implementation process.
\newblock In \textit{2020 IEEE Symposium on Security and Privacy (SP)}. IEEE.

\bibitem[{Pirch et~al.(2026)Pirch, Horlboge, Großmann, Asif, Kireev, Holz and
  Rieck}]{pirch2026securing}
\textsc{Pirch, L.}, \textsc{Horlboge, M.}, \textsc{Großmann, P.},
  \textsc{Asif, S.~M.}, \textsc{Kireev, K.}, \textsc{Holz, T.} and
  \textsc{Rieck, K.} (2026).
\newblock Toward securing {AI} agents like operating systems.
\newblock {\href{https://arxiv.org/abs/2605.14932}{\texttt{arXiv:2605.14932}}}.
\newblock 17 pages, under submission.

\bibitem[{Saltzer and Schroeder(1975)}]{saltzer1975protection}
\textsc{Saltzer, J.~H.} and \textsc{Schroeder, M.~D.} (1975).
\newblock The protection of information in computer systems.
\newblock \textit{Proceedings of the IEEE} \textbf{63} 1278--1308.

\bibitem[{Schneider(2000)}]{schneider2000enforceable}
\textsc{Schneider, F.~B.} (2000).
\newblock Enforceable security policies.
\newblock \textit{ACM Transactions on Information and System Security}
  \textbf{3} 30--50.

\bibitem[{Sedoc et~al.(2026)Sedoc, Zhang and Foster}]{sedoc2026pvd}
\textsc{Sedoc, J.~a.}, \textsc{Zhang, B.} and \textsc{Foster, D.} (2026).
\newblock Trust but verify: Prover-verifier deliberation for selective {LLM}
  prediction.
\newblock {\href{https://arxiv.org/abs/2605.25133}{\texttt{arXiv:2605.25133}}}.

\bibitem[{Seltzer et~al.(1996)Seltzer, Endo, Small and
  Smith}]{seltzer1996disaster}
\textsc{Seltzer, M.~I.}, \textsc{Endo, Y.}, \textsc{Small, C.} and
  \textsc{Smith, K.~A.} (1996).
\newblock Dealing with disaster: Surviving misbehaved kernel extensions.
\newblock In \textit{Proceedings of the Second USENIX Symposium on Operating
  Systems Design and Implementation (OSDI '96)}. USENIX Association.

\bibitem[{Shi et~al.(2025)Shi, He, Wang, Li, Wu, Guo and Song}]{shi2025progent}
\textsc{Shi, T.}, \textsc{He, J.}, \textsc{Wang, Z.}, \textsc{Li, H.},
  \textsc{Wu, L.}, \textsc{Guo, W.} and \textsc{Song, D.} (2025).
\newblock Progent: Securing {AI} agents with privilege control.
\newblock {\href{https://arxiv.org/abs/2504.11703}{\texttt{arXiv:2504.11703}}}.

\bibitem[{Steinder and Franke(2026)}]{steinder2026agentos}
\textsc{Steinder, G.} and \textsc{Franke, H.} (2026).
\newblock Towards an agent operating system - lessons from classical and cloud
  {OS}.
\newblock In \textit{2026 IEEE International Conference on Web Services
  (ICWS)}. IEEE.

\bibitem[{Sutton(2019)}]{sutton2019bitter}
\textsc{Sutton, R.} (2019).
\newblock The bitter lesson.
\newblock \url{http://www.incompleteideas.net/IncIdeas/BitterLesson.html},
  accessed 2026-10-10.

\bibitem[{Wang et~al.(2024)Wang, Chen, Yuan, Zhang, Li, Peng and
  Ji}]{wang2024codeact}
\textsc{Wang, X.}, \textsc{Chen, Y.}, \textsc{Yuan, L.}, \textsc{Zhang, Y.},
  \textsc{Li, Y.}, \textsc{Peng, H.} and \textsc{Ji, H.} (2024).
\newblock Executable code actions elicit better {LLM} agents.
\newblock In \textit{Proceedings of the 41st International Conference on
  Machine Learning (ICML)}, vol. 235 of \textit{Proceedings of Machine Learning
  Research}. PMLR.

\bibitem[{Wang et~al.(2026)Wang, Rosenberg, Michelini, Smith, Tran, Nyst,
  Malhotra, Zhou, Chen, Brennan and Neubig}]{wang2026openhandssdk}
\textsc{Wang, X.}, \textsc{Rosenberg, S.}, \textsc{Michelini, J.},
  \textsc{Smith, C.}, \textsc{Tran, H.}, \textsc{Nyst, E.}, \textsc{Malhotra,
  R.}, \textsc{Zhou, X.}, \textsc{Chen, V.}, \textsc{Brennan, R.} and
  \textsc{Neubig, G.} (2026).
\newblock The {OpenHands} software agent {SDK}: A composable and extensible
  foundation for production agents.
\newblock In \textit{Proceedings of Machine Learning and Systems (MLSys 2026)}.

\bibitem[{Watson et~al.(2010)Watson, Anderson, Laurie and
  Kennaway}]{watson2010capsicum}
\textsc{Watson, R. N.~M.}, \textsc{Anderson, J.}, \textsc{Laurie, B.} and
  \textsc{Kennaway, K.} (2010).
\newblock Capsicum: Practical capabilities for {UNIX}.
\newblock In \textit{Proceedings of the 19th USENIX Security Symposium (USENIX
  Security '10)}. USENIX Association, Washington, DC, USA.
\newblock Best Student Paper award.

\bibitem[{Wu et~al.(2024)Wu, Bansal, Zhang, Wu, Li, Zhu, Jiang, Zhang, Zhang,
  Liu, Awadallah, White, Burger and Wang}]{wu2024autogen}
\textsc{Wu, Q.}, \textsc{Bansal, G.}, \textsc{Zhang, J.}, \textsc{Wu, Y.},
  \textsc{Li, B.}, \textsc{Zhu, E.}, \textsc{Jiang, L.}, \textsc{Zhang, X.},
  \textsc{Zhang, S.}, \textsc{Liu, J.}, \textsc{Awadallah, A.~H.},
  \textsc{White, R.~W.}, \textsc{Burger, D.} and \textsc{Wang, C.} (2024).
\newblock {AutoGen}: Enabling next-gen {LLM} applications via multi-agent
  conversations.
\newblock In \textit{Conference on Language Modeling (COLM)}.

\bibitem[{Yang et~al.(2024)Yang, Jimenez, Wettig, Lieret, Yao, Narasimhan and
  Press}]{yang2024sweagent}
\textsc{Yang, J.}, \textsc{Jimenez, C.~E.}, \textsc{Wettig, A.},
  \textsc{Lieret, K.}, \textsc{Yao, S.}, \textsc{Narasimhan, K.} and
  \textsc{Press, O.} (2024).
\newblock {SWE-agent}: Agent-computer interfaces enable automated software
  engineering.
\newblock In \textit{Advances in Neural Information Processing Systems}.

\bibitem[{Zelikman et~al.(2024)Zelikman, Lorch, Mackey and
  Kalai}]{zelikman2024stop}
\textsc{Zelikman, E.}, \textsc{Lorch, E.}, \textsc{Mackey, L.} and
  \textsc{Kalai, A.~T.} (2024).
\newblock Self-taught optimizer ({STOP}): Recursively self-improving code
  generation.
\newblock In \textit{First Conference on Language Modeling (COLM)}.

\bibitem[{Zhang(2026)}]{zhang2026libos}
\textsc{Zhang, Y.} (2026).
\newblock {Agent libOS}: A runtime substrate for capability-controlled
  self-evolving {LLM} agents.
\newblock {\href{https://arxiv.org/abs/2606.03895}{\texttt{arXiv:2606.03895}}}.

\end{thebibliography}
\fi

\clearpage
\appendix
\crefalias{section}{appendix}
\crefalias{subsection}{subappendix}


\section{Formal Properties and Assumptions}
\label{app:formalization}

The implementation this appendix describes is built with Lean~4.32.0. Each name below is a graded entry of the
project's trust ledger and is found in the source under that name.
The body says what each result means; this appendix says what each result assumes and what it does not cover.

\paragraph{Formal statements.}
\begingroup\small\raggedright
\begin{description}[leftmargin=1.1em,style=nextline,font=\normalfont\itshape,nosep]
  \item[Paths and host access.]
  \lean{safePath\_no\_traversal}, \lean{cannot\_escape\_root\_via\_symlink}, \lean{admit\_refuses\_l3m}.
  \item[Grants and minting.]
  \lean{grant\_restrict\_le}, \lean{grant\_of\_record\_writes\_no\_entity\_record},
  \lean{no\_wild\_no\_wild\_effects}, \lean{mint\_reflects\_profile}, \lean{no\_self\_grant}.
  \item[Process and network checks.]
  \lean{no\_shell\_or\_interpreter}, \lean{shell\_run\_is\_card}, \lean{build\_lake\_exe\_is\_card},
  \lean{allowsUrl\_authority\_is\_trusted}, which assumes a host allowlist rather than an
  open trust floor. These are properties of the structured providers, not of every process route; the unsandboxed
  shell that \texttt{wild} grants is outside them.
  \item[Review selection and the debate.]
  \lean{ungated\_world\_reaching\_always\_reviewed}, under the default review posture;
  \lean{run\_needs\_safe\_or\_yes}, \lean{trace\_bounded}, \lean{debate\_no\_verdict\_is\_wild}.
  \item[Restoration.]
  \lean{live\_capture\_round\_trips}, \lean{live\_commit\_reset\_round\_trips}. Both take a named assumption about Git as
  an explicit hypothesis.
  \item[Sequencing and parent pins.]
  \lean{start\_once}, \lean{open\_call\_is\_last\_call}, \lean{boot\_close\_eq\_interrupt\_close},
  \lean{locked\_invariant\_under\_model\_verbs}.
\end{description}
\endgroup

\paragraph{Delegation and budgets.}
\lean{spawn\_persist\_within\_parent} relates the admission of a spawn request to the
resolved parent's grant and the row recorded for the child; it is a fact about one admission decision, not about a
whole run. \lean{descendant\_bounded\_by\_root\_grant} is the bound over the tree. It quantifies over every ledger the
agents' own writers can build, and concludes that an entity's grant lies within its root's grant expanded at the
identity of each strict ancestor (\texttt{Grant.expandOwnAll} over \texttt{ancestorChain}). The expansion adds each
ancestor's own refs and mail covers, which a child is legitimately granted; without it the bound is false at depth
one (\lean{descendant\_bounded\_by\_root\_grant\_is\_false}, with a concrete counterexample). The theorem does not bound
a descendant by every ancestor, since a regrant is checked against its issuer, and it does not extend to ledgers
containing a human's regrant, whose issuer is unbounded. The budget property \lean{global\_budget\_bounded\_by\_root}
concerns one successful \texttt{lendMany} call: for a well-formed parent lease $p$, the resulting parent $p'$ and the
returned children $c_i$,
\[
  \sum_i c_i.\mathrm{limit} + p'.\mathrm{available} \le p.\mathrm{limit}.
\]
Extending it to a whole run needs an invariant over later loans, returns and updates.

\paragraph{Sealed programs and plugins.}
\lean{shape\_admitted\_has\_no\_refused\_kind} concerns the accepted
macro-expanded syntax, and \lean{sealUsed\_admits\_only\_safe} the constants an expression names directly and their
classification in the supplied environment. The program loader runs the shape check before
elaborating the term and the expression and vocabulary checks after it, and the confinement layer keeps its
host bridge out of the vocabulary programs may name. These are implementation checks; no theorem covers the whole
loading path, and the semantics of the dependencies the runtime admits are trusted.
\lean{loaded\_plugin\_authority\_le\_cap} states that an accepted plugin has an allowed name and an admitted declared
footprint. It is an admission result, not a proof about what an arbitrary plugin does.

\paragraph{Effect evidence and credentials.}
The effect registry declares a ceiling for each class, and each declared ceiling carries evidence of one of four
kinds: irreversible, read-only frame, round-trip and host round-trip. Frame
evidence carries the names of external checks and assumptions; round-trip evidence contains a \lean{Reversible}
instance and its proof. Neither proves that the live provider implements the modelled operation.
The AWS layer defines credential resolution and signing, the profile gate and the credential fields. \lean{AwsAuthEffects} is in the program interface but has no registry row, so it takes the
conservative default ceiling and boundedness.

\paragraph{From local facts to execution.}
Each result above is a fact about a check, and it applies to a run only together with the parts of the trusted base
it does not reach. A theorem about admitted paths needs the file provider's path resolution and the file system's
behaviour; a theorem about minted tokens needs the build's constructor discipline and the providers that tools call
through; the event-loop results are about the modelled sequencer, and an external operation may need reconciliation
or an idempotent retry after a crash; and the journal round trips rest on their stated assumptions about Git and
restore only journaled effects.

\section{Related Work}
\label{app:related}

Most of the pieces L3M assembles have precedents; what is new is how they are put together. Four parts of that
combination have no close precedent. L3M is built for many agents from the start, and their authority, budget and
parentage form one tree stored in a shared Git repository. An agent extends itself with programs the harness admits
under the grant it already holds, with the program's type as its permission request. A tool's review tier is
computed from its declared effects rather than attached to each action by hand. And the theorems are about the
harness that actually runs, each graded by what its proof rests on. The paragraphs below take the neighbouring work
one area at a time: what it does, and then where L3M differs.

\paragraph{Multi-agent frameworks and agent runtimes.}
Conversation frameworks such as AutoGen~\cite{wu2024autogen} and MetaGPT~\cite{hong2023metagpt} coordinate agents
through dialogue and role prompts, and an agent in them has no authority distinct from what its prompt tells it.
Source studies of production harnesses~\cite{barbaste2026harness,liu2026claudecode} find the multi-agent case
bolted onto a single-agent core; Claude Code, for example, runs each subagent in isolation and receives only its
summary. A second line of work takes the operating system as its model. AIOS~\cite{mei2025aios} schedules model
calls, context and memory, which is resource management rather than authority. Quine~\cite{ke2026quine} makes an
agent a POSIX process and inherits isolation from the kernel, a guarantee it describes itself as incomplete.
\Citet{pirch2026securing} map classical OS defences onto agents, and \citet{steinder2026agentos} list the
primitives an Agent-OS still lacks, among them an abstraction for irreversible effects and one for messages between
agents. The closest system to L3M is Agent libOS~\cite{zhang2026libos}, which states the invariant L3M also rests
on: an agent ``may change what it can ask for, but it cannot thereby change what it is authorized to affect''. It
admits an operation only when process identity, task authority, a typed capability, policy or human approval, and
budget all agree; forked children receive attenuated authority; just-in-time tools run behind a single syscall
entry point; and a typed Git provider, pinned to one repository, uses compare-and-swap state tokens. Elsewhere, where
Git or an event log appears, it checkpoints a single agent: the OpenHands SDK~\cite{wang2026openhandssdk}
event-sources one conversation and names multi-agent coordination as undesigned, and
SWE-Search~\cite{antoniades2025swesearch} commits each repository state so that its search can backtrack.

In L3M, Git is not one provider among others; it holds all of the state. An agent is a column of refs in a shared
bare repository --- its log, inbox, grants, claims and persona branch --- together with a worktree it is confined to.
A child's name extends its parent's, and so does its directory; parentage is immutable, a grant that would widen the
issuer's own is refused rather than clamped, and budget is lent down the tree and conserved. Mail is an append to
the recipient's inbox ref. Inboxes merge by set union, which we prove, so delivery needs no recipient awake and
crosses machines by fetch and push. Each log has one writer, enforced by compare-and-swap on its ref, so a second
writer is a conflict rather than an interleaving --- the durable-workflow pattern, applied to an agent whose largest
activity is a model call. A persona is a ref that many agents branch from, and its authority is a ceiling that can
only subtract. Agents sleep on bounded watches and are woken by facts in the repository, and an ancestor can pin a
watch on a child as an obligation the child cannot cancel. Because the permission tree, the ref tree and the
directory tree are the same tree, a bound checked at each spawn constrains the whole subtree beneath it, and
cleaning up after an agent is collecting that subtree. Two further differences separate L3M from libOS. Where libOS admits an operation
by a check at a Python syscall boundary, and evidences it with hashed reports from scripted runs, L3M bounds a
generated program by how its Lean term elaborates, and its admission rules are theorems. And where libOS asks for
approval per operation, L3M ranks effects by whether they can be undone and spends review only where they cannot.

\paragraph{Capabilities and least authority.}
L3M's authority model is the classical one: capabilities as unforgeable references~\cite{dennis1966semantics}, and
least privilege with complete mediation~\cite{saltzer1975protection}. Prompt injection is the confused
deputy~\cite{hardy1988confused} with a model as the deputy. \Citet{miller2003paradigm} separate what a subject is
permitted to do from what it can bring about, and ``capability grows; authority does not'' is that distinction.
Joe-E~\cite{mettler2010joee} removes ambient authority from a language and Capsicum~\cite{watson2010capsicum} adds
capabilities to an operating system; macaroons~\cite{birgisson2014macaroons} attenuate a credential at every hop,
as L3M's grants narrow at every spawn. \Citet{devriese2016parametricity} prove that bounding the capabilities code
holds bounds its effects, which is what L3M's effect classes provide: a tool body written over an abstract monad
cannot name \texttt{IO}. What is different in L3M is the code under the discipline. It is written by the model while
the system runs, and its bound is checked by Lean's elaborator when the program is admitted, not by a runtime
monitor.

\paragraph{Safe extension and self-extending agents.}
Admitting untrusted extensions to a trusted core is an old systems problem. SPIN~\cite{bershad1995spin} relies on
a type-safe language; VINO~\cite{seltzer1996disaster} wraps each graft in a transaction that is undone on abort;
proof-carrying code~\cite{necula1997pcc} moves trust from the producer of code to a small checker; and
Singularity~\cite{hunt2007singularity} seals processes so that no code can be loaded into them after they start.
Agents that write code have no such bound. Models make their own tools~\cite{cai2024latm} and act by writing
code~\cite{wang2024codeact}, but nothing states what authority that code holds, and self-improving scaffolds have
been observed disabling their own sandbox~\cite{zelikman2024stop}. In L3M the sealed program is the extension and
Lean is the checker. A program passes three gates: a seal that refuses unsafe constants, a type under which
undeclared effects are unreachable, and a surface check that offers it only if its declared footprint fits the
grant. Once admitted, a program does the looping that the model would otherwise do one call at a time, and it can
delegate to children it drives to completion, all under that same grant (\Cref{fig:workflow}). Reversibility is graded per effect, rather than resting on one undo log for everything.

\paragraph{Enforcement for LLM agents.}
Indirect prompt injection~\cite{greshake2023injection} and the AgentDojo benchmark~\cite{debenedetti2024agentdojo}
set out the threat. The defences closest to L3M work inside a fixed agent loop. CaMeL~\cite{debenedetti2025camel}
and FIDES~\cite{costa2025fides} label data and check its flows through a planner; Progent~\cite{shi2025progent}
narrows tool policies monotonically; and LLMbda~\cite{garby2026llmbda} mechanises an information-flow calculus in
Lean~4, proves noninterference, and runs the verified interpreter as the harness. That makes LLMbda the nearest
work on proving properties of the code that runs, though its tools are modelled inside the program.
L3M differs from these designs first in what it bounds: the programs the model generates, by their type, acting
through real file-system, Git and process effects. Every proof that depends on how the outside world behaves names
that dependency as an explicit assumption. LLMbda proves one strong property of a calculus the size of a harness;
L3M proves the properties each mechanism needs, of the full harness. L3M does not yet bound information flow: its
design for doing so (\Cref{app:infoflow}) is not in the implemented version. That design places its labels
differently. Data at rest carries a label, an artifact carries its own bound tied to the hash of its content, and every
edge between agents charges the sender and joins the sender's taint into the receiver, so the bound follows data
across agents. The meter is at the read, the only place attribution is possible, and what leaks of a kind is bounded
by both what entered and what was sent. That makes the information-flow bound a safety property, which by
\citet{schneider2000enforceable} is the class an execution monitor can enforce; the per-call gate where L3M already
meters egress would be such a monitor, and the design claims nothing a monitor cannot enforce. Beneath any labels, the
capability bound holds today: an injected instruction can reach only what the agent was granted.

\paragraph{Reversibility, review and the human.}
Magentic-UI~\cite{mozannar2025magenticui} sends every action through a guard that developer heuristics have
labelled always, maybe or never irreversible, and the Agent-OS proposal~\cite{steinder2026agentos} asks for
operations to be classed as reversible, compensable or irreversible. L3M computes a tool's tier from its declared
footprint instead, so no tool is labelled by hand. Reversibility is stated once per effect class, as a ceiling
on how far that class's effects can be undone: proved as a round trip where it can be --- a journaled file edit, a
local Git ref, an evaluation --- with any dependence on Git's behaviour named as an assumption, and labelled where
it cannot. A theorem refuses any tool that claims more reversibility than its footprint allows.

Users approve about 93\% of Claude Code's permission prompts~\cite{liu2026claudecode}, so approving every call is
not oversight. L3M spends the human's attention only past the reversible margin, and there review is a deliberation,
in the line of AI debate~\cite{irving2018debate}, prover--verifier games~\cite{anil2021pvg,browncohen2025avoiding},
and control protocols in which a trusted monitor watches an untrusted model~\cite{greenblatt2024control}. From
Prover--Verifier Decoding~\cite{sedoc2026pvd} we take the three verdicts, the bounded exchange, cross-family
pairing, and a proposer who must supply a justification. We do not take its soundness bounds or its
selective-prediction framing: L3M always reaches a verdict, and abstention hands the decision to the human. The
reviewer is a way to spend attention cheaply, not a security boundary; the capability discipline holds without it.
The human's place follows mixed-initiative design~\cite{horvitz1999mixedinitiative}, and heeds Bainbridge's warning
that automation leaves the operator only the hardest cases, and without practice at
them~\cite{bainbridge1983ironies}. A card therefore shows the effect and the reviewer's reason rather than a shell
string, and a card that is almost always approved is re-aimed or removed. The aim is Licklider's division of
labour~\cite{licklider1960symbiosis}, applied to agents.

\paragraph{Verified systems.}
seL4~\cite{klein2009sel4} proves a kernel's implementation correct against its specification,
CompCert~\cite{leroy2009compcert} verifies a compiler, and CHERI~\cite{nienhuis2020cheri} shows what formalising a
prose design can uncover. L3M does not aim for one whole-system theorem. It is
written in Lean~4~\cite{demoura2021lean4}, so each theorem it does have is about the code that runs, and it is
formalized incrementally: every claim is graded by what its proof uses, and a levelized source tree fixes what each
claim may rest on (\Cref{app:formalization}).

\section{Bounding Information: An Extension}
\label{app:infoflow}

Capabilities bound which edges exist, and the budgets of the body bound how many calls, bytes and wakes cross them.
Neither bounds how much of a \emph{kind} of information crosses an edge that exists: an agent that may read payroll
and may fetch a public page may also encode the one into the other. This appendix records the design that closes
that gap. It is a design, and not in the version of L3M implemented and described in the body of the paper.

\paragraph{The model.} Data at rest carries a \emph{label}, a kind of information, drawn from a lattice with two
built-in points: $\mathsf{Public}$ at the bottom and $\mathsf{NonPublic}$ at the top, with user-declared kinds
(\texttt{payroll}, \texttt{pii}) between them. Two different kinds join to $\mathsf{NonPublic}$, and nothing joins
downward. A \emph{metered edge} is any boundary across which bytes enter or leave a principal's state: a file read,
a fetch, a tool result, a stored artifact read back later, a message to or from another agent. Each agent keeps a
\emph{taint set}, the join of every label it has read, and for each label $K$ a monotone accumulator
$\mathrm{ingress}[K]$ of the bytes of $K$ that have entered its state. The accumulator only grows and never refuses;
a \emph{limit} is a separate gate that reads it, so accounting soundness never depends on policy. The ledger is a
field of the agent's pure state, changed only by the fold of \Cref{sec:log}, so facts about it are theorems rather
than runtime hopes.

\paragraph{Why the meter is at the read.} Model an agent as a chain
$\text{Secret}_K \to \text{State} \to \text{Output}$. By the data-processing inequality,
\[
  I(\text{Secret}_K ; \text{Output}) \;\le\; I(\text{Secret}_K ; \text{State}) \;\le\;
  \text{the bytes of $K$ that entered the state}.
\]
Charging $K$ when it is read therefore bounds what can leave, however much the agent outputs; and the read is the
only place attribution is possible, because once the model has transformed its state no output byte still wears a
label. The inequality is a statement about information, not about L3M, so it enters the design the way Git's
behaviour enters the body: as a named assumption that every dependent claim takes as a hypothesis.

\begin{proposition}[Leakage bound, under the data-processing inequality]
\label{prop:leak}
Keep per label a second accumulator $\mathrm{egress}[K]$, the bytes sent while $K$-tainted. Then for every $K$,
$\mathrm{leak}(K) \le \min(\mathrm{ingress}[K],\ \mathrm{egress}[K])$, with the minimum taken once over the
accumulated totals, never per request.
\end{proposition}

\noindent The order matters: an agent holding 64 bytes of $\mathsf{NonPublic}$ that sends two 50-byte requests has
leaked at most $\min(64, 100) = 64$, not $\min(50,64) + \min(50,64) = 100$. An outbound request is charged at its
full width to every label the sender holds, because a secret can be smuggled into any byte of it; the bound then
says the agent cannot leak more of a kind than ever entered.

\paragraph{Completeness is structural.} The bound holds only if no edge escapes the meter. Every edge is one of
three things: a \emph{charging} edge, labelled by a rule or by the artifact's own stamp; a \emph{framework} edge,
whose bytes originate in the trusted base (the system prompt, the tool catalogue, a clock) and which is uncounted
because it has no labelable channel; or a world edge that matches no rule, which charges $\mathsf{NonPublic}$,
never nothing. In L3M every ingress edge is an effect class closed by one provider (\Cref{sec:tokens}), so ``every
edge is metered'' is ``every provider charges at its boundary'': a census over the enumerable trusted base, not a
scan of the tree, and a new tool is metered the moment it is registered, because the dispatcher, not the tool's
author, does the accounting.

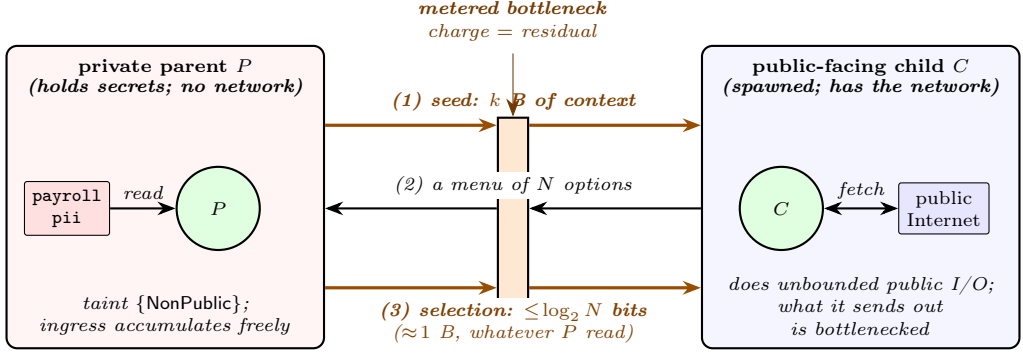
\begin{figure}[t]
\centering
\begin{tikzpicture}[>={Stealth[length=2.4mm]}, font=\footnotesize,
  env/.style={draw, rounded corners=4pt, thick, inner sep=6pt, align=center},
  hdr/.style={font=\scriptsize\bfseries, align=center},
  ag/.style={draw, circle, thick, fill=green!12, minimum size=11mm, align=center, font=\scriptsize\bfseries},
  data/.style={draw, rounded corners=1pt, fill=red!12, inner sep=3pt, align=center, font=\scriptsize},
  net/.style={draw, rounded corners=1pt, fill=blue!10, inner sep=3pt, align=center, font=\scriptsize},
  note/.style={font=\scriptsize\itshape, align=center},
  msg/.style={note, fill=white, inner sep=1.5pt}]
  \node[env, fill=red!4,  minimum width=42mm, minimum height=40mm] (privbox) at (0,0)   {};
  \node[env, fill=blue!4, minimum width=42mm, minimum height=40mm] (pubbox)  at (9.2,0) {};
  \node[hdr, anchor=north] at ([yshift=-2pt]privbox.north)
    {private parent $P$\\[-1pt] \textup{\itshape(holds secrets; no network)}};
  \node[data] (secret) at (-1.3,-0.15) {\texttt{payroll}\\ \texttt{pii}};
  \node[ag]   (par)    at (0.7,-0.15)  {$P$};
  \draw[->, thick] (secret) -- node[note, above]{read} (par);
  \node[note, anchor=south, text width=40mm] at ([yshift=2pt]privbox.south)
    {taint $\{\mathsf{NonPublic}\}$;\\ ingress accumulates freely};
  \node[hdr, anchor=north] at ([yshift=-2pt]pubbox.north)
    {public-facing child $C$\\[-1pt] \textup{\itshape(spawned; has the network)}};
  \node[ag]  (chi)  at (8.15,-0.15) {$C$};
  \node[net] (net)  at (10.3,-0.15) {public\\ Internet};
  \draw[<->, thick] (chi) -- node[note, above]{fetch} (net);
  \node[note, anchor=south, text width=40mm] at ([yshift=2pt]pubbox.south)
    {does unbounded public I/O;\\ what it sends out is bottlenecked};
  \node[draw, thick, fill=orange!18, minimum width=4mm, minimum height=24mm, inner sep=0pt] (gate) at (4.6,-0.15) {};
  \draw[->, very thick, orange!60!black] (privbox.east|-0,0.95) -- (gate.west|-0,0.95);
  \draw[->, very thick, orange!60!black] (gate.east|-0,0.95) -- (pubbox.west|-0,0.95);
  \node[msg, anchor=south, text=orange!50!black] at (4.6,1.08) {\textbf{(1) seed: $k$ B of context}};
  \draw[->, thick] (pubbox.west|-0,-0.15) -- (gate.east|-0,-0.15);
  \draw[->, thick] (gate.west|-0,-0.15) -- (privbox.east|-0,-0.15);
  \node[msg, anchor=south] at (4.6,-0.03) {(2) a menu of $N$ options};
  \draw[->, very thick, orange!60!black] (privbox.east|-0,-1.2) -- (gate.west|-0,-1.2);
  \draw[->, very thick, orange!60!black] (gate.east|-0,-1.2) -- (pubbox.west|-0,-1.2);
  \node[msg, anchor=north, text=orange!50!black] at (4.6,-1.33)
    {\textbf{(3) selection: $\le\!\log_2 N$ bits}\\ ($\approx\!1$ B, whatever $P$ read)};
  \node[note, anchor=south, text width=30mm, orange!45!black] at (4.6,1.95)
    {\textbf{metered bottleneck}\\ charge $=$ residual};
  \draw[orange!45!black, ->] (4.6,1.9) -- (gate.north);
\end{tikzpicture}
\caption{The bottleneck. A private parent $P$ holds the secrets and has no network; its child $C$ has the network and
no secrets. Every message across the boundary is a metered edge, and the charged direction is private to public
(orange). $P$ seeds $C$ with $k$ bytes (1); $C$ works publicly and returns a menu of $N$ options (2); $P$ answers
with a selection (3), so at most $\log_2 N$ bits leave the private side however much $P$ read. In L3M this is a
spawn pattern over the grants of \Cref{sec:ex-split}.}
\label{fig:bottleneck}
\end{figure}

\paragraph{Information, not bytes.} What crosses an edge is charged at a computable upper bound on the information
it carries, $I(\text{Secret}_K; M) \le H(M) \le |\mathrm{encode}(M)|$, so any lossless encoding is a sound meter
and a tighter one is still sound. A fixed compressor makes boilerplate cheap and novelty expensive. The best
compressor, though, is a capable agent on the receiving side: what $B$ gains from a message is only what $B$ could
not reconstruct by interacting with the world itself, so the sound charge is that residual, and a sender that has
read a great deal ships the smallest thing the receiver cannot rebuild. At the limit the private side ships a
\emph{choice} (\Cref{fig:bottleneck}): a selection among $N$ options carries at most $\log_2 N$ bits, one byte for
$N \le 256$, whatever was read to make it. The bound needs the recipient to act on the index into a menu it
understands independently; secret-derived contents of the options are charged as ordinary bytes.

\paragraph{Artifacts and other agents.} Information parked on disk and read back later, and information passed
between agents, reduce to the same rule: an artifact carries its own bound, and every boundary it crosses is one
metered edge. A stamp binds a bound to the \emph{content hash} of the bytes, never to a path, and is discharged at
read time against the bytes actually present: a hit charges the stamped bound, a miss charges the full size at
$\mathsf{NonPublic}$. An adversary who swaps the bytes changes the hash and lands on the floor, so there is no
time-of-check window. The stamp records the taint that \emph{flowed into} the artifact, not its producer's balance:
a producer re-reading its own file pays nothing, since the file can hold only what it already paid for, and a
disk-to-disk fetch is stamped at the width of the request that caused it, by the same inequality applied to
request and response. Between agents, every edge charges the sender and joins the sender's taint into the receiver,
parent to child and peer to peer alike, so the bound composes over any topology and sideways flow cannot bypass it.
A child's return is charged to the parent at $\min(|R|,\ \mathrm{ingress}_C[K] - \mathrm{from}_P[K])$: not the raw
size of $R$, which ignores the cap on the return channel, and not the child's whole intake, which double-counts what
the parent seeded.

\end{document}